%% file: 0_main.tex
\documentclass[final,5p,times,twocolumn,number]{elsarticle}
\usepackage[switch]{lineno}
\usepackage{textgreek}
\usepackage[utf8]{inputenc}
\usepackage{graphicx}

\usepackage[table]{xcolor}

\usepackage{xurl}
\usepackage{amssymb}
\usepackage{amsmath,amsfonts,amsthm,bm}
\usepackage{bigints}
\usepackage{subfigure}
\usepackage{pgfplots}
\usepackage{comment}
\usepackage{enumitem}
\usepackage{pifont}
\usepackage{booktabs}
\usepackage{gensymb}
\usepackage{float}

\pgfplotsset{compat=1.7}
\usepgfplotslibrary{colorbrewer}
\usepgfplotslibrary{colormaps}

\graphicspath{{./Figures/}}

\usepackage{titlesec}
\titleformat{\paragraph}{\normalfont\normalsize\bfseries}{\theparagraph}{1em}{}
\titlespacing*{\paragraph}{0pt}{3pt}{3pt}

\usepackage{setspace}
\biboptions{sort&compress}

\usepackage{caption}
\makeatletter
\def\ps@pprintTitle{%
 \let\@oddhead\@empty
 \let\@evenhead\@empty
 \def\@oddfoot{}%
 \let\@evenfoot\@oddfoot}
\makeatother

\usepackage{threeparttable}

\usepackage{hyperref}
\hypersetup{colorlinks=true, citecolor=blue, linkcolor=blue, filecolor=blue, urlcolor=blue,}

\begin{document}


\begin{frontmatter}


\title{Learning Diffusion Policies for Robotic Manipulation of Timber Joinery under Fabrication Uncertainty}


\author[add1]{Salma Mozaffari\corref{cor1}}
\ead{salma.mozaffari@princeton.edu}

\author[add1]{Daniel Ruan\corref{cor1}}
\ead{daniel.ruan@princeton.edu}

\author[add2]{William van den Bogert}
\ead{willvdb@umich.edu}

\author[add2]{Nima Fazeli}
\ead{nfz@umich.edu}

\author[add1]{Sigrid Adriaenssens}
\ead{sadriaen@princeton.edu}

\author[add1]{Arash Adel\corref{cor2}}
\ead{arash.adel@princeton.edu}

\address[add1]{Princeton University, Princeton, NJ 08544, USA}
\address[add2]{University of Michigan, Ann Arbor, MI 48109, USA}
\cortext[cor1]{Equal contribution.}
\cortext[cor2]{Corresponding author.}


\begin{abstract}

Fabrication uncertainty arising from tolerance accumulation, material imperfection, and positioning errors remains a critical barrier to automated robotic assembly in construction, particularly for contact-rich manipulation tasks under minimal geometric clearance. This paper investigates the deployment of diffusion policy learning on construction-scale industrial robots to enable robust, high-precision assembly under such uncertainty, using tight-clearance mortise and tenon timber joinery as a representative case study. Sensory–motor diffusion policies are trained using teleoperated demonstrations collected from an industrial robotic workcell equipped with force/torque sensing. A two-phase experimental study evaluates baseline performance and robustness under randomized positional perturbations up to 10 mm, far exceeding the joint clearance. The best-performing policy achieved 100\% success under nominal conditions and 75\% average success under uncertainty. These results suggest that diffusion policies can improve robustness to fabrication-induced misalignment, representing a step toward reliable robotic assembly in construction under tight tolerances.

\end{abstract}


\end{frontmatter}

\input{1_introduction.tex}

\input{2_related_work.tex}

\input{3_methods.tex}

\input{4_experiments}

\input{5_results.tex}

\input{6_conclusion.tex}

\section*{Acknowledgments}
This research was supported by the National Science Foundation (NSF, Award No. 2122271), Princeton Catalysis Initiative, and the School of Architecture (SoA) at Princeton University. The authors would like to thank Ruxin Xie, Zhengyi Chen, and Roman Ibrahimov at the Adel Research Group (ARG) and SoA for their invaluable support in hardware development and data collection.

\section*{Data availability statement}
The data used in this study are available upon request. 

\bibliographystyle{IEEEtran} 
\bibliography{bibliography.bib}

\end{document}

%% file: 1_introduction.tex
\section{Introduction}\label{introduction}

Construction plays a critical role in the global economy but continues to face long-standing challenges, including stagnant productivity growth, shortages of skilled labor, and persistent health and safety concerns~\cite{delgado2019, wei2023}. The industry often requires workers to perform repetitive, physically demanding tasks, such as lifting and positioning heavy components, which can lead to chronic injuries and reduced long-term workforce capacity~\cite{laukkanen1999, arndt2005}. Automation, and in particular robotic assembly, offers the potential to address these issues by enhancing precision, increasing efficiency, enabling mass customization, and reducing the physical burden on workers~\cite{bock2015, musarat2024}. 

Recent advances in construction robotics have increasingly leveraged industrial manipulators originally developed for automation in factories~\cite{chen2025}. These systems play a central role in large-scale assembly due to their high payload capacity and long reach, enabling the manipulation of heavy structural components and customized geometries that would be difficult to achieve manually. However, unlike structured factory environments, construction sites are inherently uncertain. Translating the capabilities of industrial robots from manufacturing to these uncertain conditions, characterized by fabrication inaccuracies, material imperfections, and dynamic environments, remains a significant challenge~\cite{adel2024, delgado2019, musarat2024}. Under such conditions, pre-programmed trajectories executed without feedback often result in misalignment, collision, and task failure, exposing the limits of purely feedforward control in construction contexts.

The momentum in construction robotics research is especially evident in recent work on robotic timber assembly~\cite{willmann2016, apolinarska2018, adel2018, adel2020, graser2021, chai2022, lauer2023, ruan2023, adel2026}, where many workflows rely on planar face-to-face or butt joints, whose geometric simplicity allows the use of pre-programmed trajectories for successful assembly. In contrast, the manipulation of timber joinery such as mortise and tenon, lap, or scarf joints requires contact-dominated insertion under minimal geometric clearance and demanding high precision and careful regulation of contact forces~\cite{apolinarska2018, leung2021}. In these scenarios, even minor positional deviations can produce jamming, stick–slip behavior, or material damage if insertion is not coupled with real-time control and robust error recovery~\cite{apolinarska2021, kramberger2022}. As a result, these tasks are often delegated to skilled human workers, who intervene to clamp, hammer, and correct misalignments, compensating for fabrication inaccuracies and material imperfections~\cite{yang2024}. Such manual corrections interrupt digital workflows, limit scalability, and constrain the adoption of fully automated processes. Furthermore, it increases reliance on a shrinking pool of skilled labor and entails physically demanding work that contributes to fatigue and injury.

\begin{figure*}
    \centering
    \includegraphics[width=0.8\linewidth]{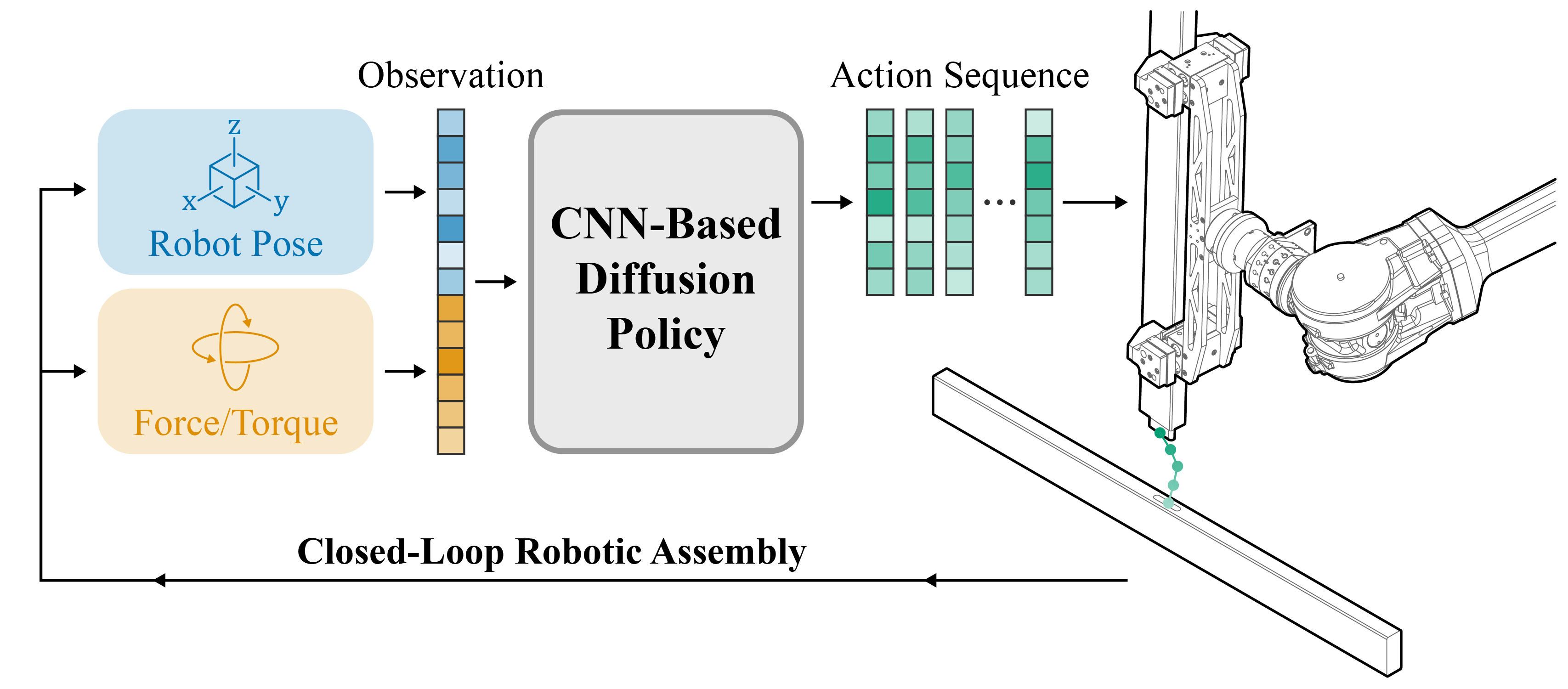}
    \caption{Overview of our method. A CNN-based diffusion policy is trained conditional on end effector pose (i.e., position and orientation) and F/T observations to predict sequences of robot actions.}
    \label{fig:overview}
\end{figure*}

Timber joinery is a centuries-old construction technique of geometrically interlocking timber components to create durable structural connections with minimal reliance on fasteners or adhesives~\cite{benson1981}. These joinery systems are still widely used in Asian construction and are being revisited in Europe and North America~\cite{structurecraft, shinohara}. These joints are valued for their enhanced structural performance, rotational stiffness, and resilience under seismic and fire conditions~\cite{fang2023}. However, their complex geometry and contact-dominated assembly process make them particularly sensitive to fabrication inaccuracies and material imperfections, including cut length and angle deviations, component shifting during picking, placing, and fastening, and dimensional changes due to hygrothermal effects~\cite{adel2024}. When joint clearances approach the sub-millimeter scale, tolerance accumulation across assembly steps and geometric variability can easily exceed allowable insertion margins, rendering purely feedforward execution unreliable in practice. Despite the significance of this challenge and the advanced structural performance of these joints, their automated assembly remains nascent in robotic workflows within the Architecture, Engineering, and Construction (AEC) domain. Overcoming these limitations requires feedback-driven, closed-loop planning and control methods capable of adapting to the unpredictable and evolving contact interactions. 

Learning-based methods have increasingly been adopted as alternatives to traditional model-based control methods for closed-loop robot planning and control, particularly in tasks involving complex contact dynamics and uncertainty. Sensory--motor policy learning methods integrate sensor observations, planning, and control into a single end-to-end mapping, originally implemented using deep convolutional neural networks (CNNs)~\cite{levine2016}. These learned control policies map raw observations, such as images captured by a camera and force feedback, directly to motor commands or robot actions using various learning-based algorithms~\cite{finn2017, levine2018, kalashnikov2018, brohan2023}. Among these methods, diffusion policy learning via behavior cloning~\cite{chi2024a} has demonstrated strong performance across a range of dexterous manipulation scenarios~\cite{chi2024b, hou2025, yang2025, huaijiang2025}. However, despite these advances, the systematic deployment and evaluation of these sensory–motor policies on industrial manipulators remain limited. Industrial systems introduce additional constraints, including high payloads, large inertia, limited intrinsic compliance, and safety-certified control interfaces, that complicate stable, real-time manipulation~\cite{kommey2025, ruan2026}. These challenges are particularly pronounced in contact-rich assembly tasks, where geometric imperfections and fabrication inaccuracies demand robust closed-loop control.

This paper addresses contact-rich robotic manipulation under \emph{fabrication uncertainty}, defined here as the combined influence of tolerance accumulation across assembly steps and geometric imprecision arising from material imperfections, both of which introduce positional misalignment in tight-fitting timber joints and hinder precise and robust robotic assembly. Unlike many recent studies primarily validated on tabletop robotic platforms, we systematically deploy and evaluate diffusion policy learning on construction-scale industrial manipulators operating under real-world hardware and control constraints. Our sensory--motor diffusion policies are trained on robot end effector pose and force/torque (F/T) data from teleoperated demonstrations, enabling precise assembly of a mortise and tenon joint as a representative contact-rich manipulation case study (Fig.~\ref{fig:overview}).

\subsection{Objectives and contributions}

This study pursues three primary objectives: (1) Evaluate the applicability of diffusion policies for contact-rich robotic assembly in a construction-scale case study of a mortise and tenon joint with sub-millimeter clearance; (2) Systematically assess policy robustness under fabrication uncertainty, modeled as randomized positional perturbations of the mortise; and (3) Analyze the effects of training and inference parameters, sensing modalities, and demonstration count on robust performance on large-scale industrial robotic arms.

This work presents one of the first systematic investigations of diffusion policy learning in the AEC domain for contact-rich assembly under fabrication uncertainty at construction-scale. We evaluate policy robustness under fabrication uncertainty and provide a structured assessment of performance across varying positional perturbations, sensing modalities, and training dataset size. Importantly, we investigate deployment on industrial robotic arms, where applying end-to-end sensory–motor policy learning remains challenging due to hardware and control constraints such as high payloads, limited intrinsic compliance, restricted access to low-level control, and conservative safety-certified interfaces.

By enabling autonomous high-precision assembly without reliance on human intervention, the proposed framework addresses physically demanding, craftsmanship-intensive tasks and advances the feasibility of automated construction workflows. The findings provide evidence applicable to a broader range of contact-rich manipulation tasks in construction, including complex timber joints, pipe fitting, and light metal framing. Ultimately, this work advances robotic construction under uncertainty by bridging state-of-the-art sensory--motor policy learning methods with real-world AEC workflows and provides practical insights that can accelerate adoption across the field.

%% file: 2_related_work.tex
\section{Related work}\label{related_work}

In this section, we first review prior work on robotic timber assembly under uncertainty, including contact-rich manipulation of timber joints. Next, we summarize recent advances in sensory–motor policy learning for robot planning and control. Finally, we position existing robot learning efforts in construction tasks and motivate the present study.
 
\subsection{Robotic timber assembly}

In construction with discrete elements, such as brickwork and timber or steel framing, uncertainty remains a fundamental challenge to achieving precise and robust robotic assembly. In robotic timber assembly, which serves as the case study in this paper, these uncertainties arise from material imperfections (e.g., dimensional deviations, deformation, shrinkage/expansion due to moisture) and fabrication inaccuracies (e.g., cut length and angle errors, robot pose or grasp pose deviation). If unaccounted for, these perturbations cause error accumulation during assembly, leading to misalignment, part collision, and task failure when trajectories are executed directly from as-planned digital models ~\cite{adel2024}. Prior research in robotic timber assembly has explored various strategies to address these uncertainties, including computational modeling of tolerance propagation~\cite{gandia2022}, adaptive, feedback-driven methods~\cite{helm2016, adel2024, cote2024, xie2026}, and external tracking systems for end effector pose correction in multi-robot assembly workflows~\cite{stadelmann2019, adel2020}. However, the successful and precise execution of timber joints often still depends on human intervention to correct deviations during nailing, gluing, and insertions~\cite{willmann2016, adel2018, adel2020, helmreich2022}. The challenges of uncertainty become even more pronounced in contact-sensitive assembly tasks, such as those in traditional timber joinery, where success depends not only on geometric precision but also on controlling excessive force interactions arising from friction and tight tolerances. In such scenarios, even minor deviations introduced during fabrication or caused by material imperfections can easily lead to misalignment, resulting in jamming, stick-slip behavior, and material degradation~\cite{apolinarska2021, kramberger2022}.

\subsection{Contact-rich manipulation of timber joints}\label{crm_construction}

Handling force interactions in contact-rich robotic manipulation is traditionally addressed through model-based impedance or admittance control laws~\cite{albu2007, suarez2018}, learning-based control policies~\cite{vecerik2019, schoettler2020}, or hybrid approaches combining both~\cite{johannsmeier2019, hou2025}. In the context of assembly with timber joinery, a few studies have explored learning-based methods. Apolinarska et al.\ applied reinforcement learning for the motion control of an industrial robot to assemble single- and double-lap joints using robot pose and force/torque data from a wrist-mounted sensor~\cite{apolinarska2021}. For the single-lap joints, they also investigated policy robustness by introducing angular and linear offsets to the initial robot pose while grasping the top piece. Their results showed strong performance in zero-offset cases and revealed challenges of policy adaptation when angular and linear offsets, as well as angled or double insertions, were introduced. The findings underscored the importance of robust policy learning for handling geometric variations and positional deviations in contact-rich construction assembly tasks. Kramberger et al.\ also proposed a learning-by-demonstration method that integrated a compliance controller into Cartesian-space dynamic movement primitives for perpendicular single-lap joint insertion; however, they relied on significantly loose tolerances~\cite{kramberger2022}. 

Other studies have explored semi-structured or open-loop methods for timber joinery assembly; Robeller et al.\ introduced a custom end effector that generated vibration and combined it with a manually applied mallet-driven pulse force to facilitate the connection of wood panels using dovetail joints~\cite{robeller2017}. Leung et al.\ demonstrated the use of an industrial robot, combined with manually placed distributed robotic clamps, to assemble tight-fitting half-lap joints~\cite{leung2021}. During their experiments, they observed challenges such as misalignments and collisions due to fabrication inaccuracies and deviations between digital and as-built models, which are common in construction-scale robotic applications. Finally, Rogeau employed visual feedback and fiducial markers for the assembly of wood panels with multiple mortise and tenon joints and reported successful full insertions in about 50\% of their test cases across various joint configurations such as mortise tightness and tenon chamfer angles~\cite{rogeau2023}.

\subsection{Sensory--motor policy learning}

Model-based control methods, which require explicit knowledge of system dynamics, can be particularly difficult to apply when handling complex contact dynamics, such as those found in multi-body and multi-surface interactions or in the manipulation of deformable objects. Sensory--motor policy learning methods have increasingly been adopted as effective alternatives to bypass the need for accurate system dynamics by learning control policies that are conditioned on sensor observations, such as images and force feedback, to produce robot actions~\cite{levine2016, finn2017, levine2018, kalashnikov2018, brohan2023}. Among these learning methods, behavior cloning is widely adopted, as it replaces task-specific robot programming with human demonstrations, typically collected through teleoperation~\cite{seo2023, wang2023, shaw2023}. Policy learning from human demonstrations can be perceived as a sequenced supervised learning problem that maps a history of observations to the sequence of actions (robot commands). With access to sufficient demonstration data, behavior cloning has shown promise in tasks involving challenging conditions such as deformable objects and bimanual coordination~\cite{zhao2023, zhao2025}.

Diffusion policies~\cite{chi2024a, ze2024, zhao2025} represent a recent class of behavior cloning methods built upon Denoising Diffusion Probabilistic Models (DDPMs)~\cite{ho2020} and Denoising Diffusion Implicit Models (DDIMs)~\cite{song2022}. DDPMs are generative models originally developed for high-quality image synthesis. These models learn to predict and iteratively remove noise from a corrupted sample, gradually transforming it into a sample from the target data distribution. DDIMs extend this framework by formulating a non-Markovian denoising process, which allows the number of steps used during inference to be reduced without changing the training procedure, thereby enabling significantly faster sampling. In diffusion policy via robot action diffusion~\cite{chi2024a}, the denoising process generates a sequence of robot trajectories at each inference step. This method is known to capture multimodal action distributions, meaning it can represent multiple distinct yet equally valid action sequences for completing a task, as often occurs in human demonstrations. The approach also maintains temporal coherence, meaning the policy predicts an entire sequence of future actions jointly rather than one step at a time. This joint prediction ensures that consecutive actions remain consistent rather than switching erratically between different predicted single actions, which can lead to unstable or jittery motion. Moreover, the method avoids the need for action discretization (breaking continuous movements into a fixed set of bins), or negative sampling (generating incorrect actions during training to guide learning)~\cite{chi2024a}. This removes common limitations of explicit behavior cloning methods~\cite{florence2020, shafiullah2022}, which can make it difficult to achieve high-precision actions. Diffusion policies also offer improved training stability compared to implicit behavior cloning methods~\cite{florence2022, jarrett2020}. In addition, they have demonstrated robustness to perturbations, visual occlusions, and idle actions in some manipulation tasks~\cite{chi2024a}.

\subsection{Robot learning in construction}

Planning and control through contact remains a major challenge due to the inherent complexity of contact dynamics~\cite{yang2025}. While diffusion policies have achieved notable success in a range of dexterous manipulation tasks, their effectiveness in highly contact-rich scenarios is still under active investigation~\cite{hou2025, wu2025, zhou2026}. In particular, the deployment of these end-to-end sensory--motor policies on industrial manipulators for dexterous manipulation tasks has remained underexplored, likely due to compounded challenges arising from hardware constraints, high payloads, limited intrinsic compliance and force control, conservative safety-certified control interfaces, and restricted access to low-level control~\cite{kommey2025, ruan2026, delgado2022}.

A growing body of work has investigated robot learning for construction tasks, most commonly through imitation learning and reinforcement learning~\cite{sun2026, huang2026, duan2024, duan2025, yu2024, li2023, apolinarska2021}. Notably, most of these studies were conducted on lightweight collaborative or mobile manipulators operating under relatively compliant hardware conditions. Among these, Sun et al.\ investigated rebar slot insertion and rebar tying using a mobile manipulator, combining visual servoing for coarse positioning with imitation learning for precise execution~\cite{sun2026}. The rebar insertion task required precise alignment of the stirrup with three slots prior to insertion, with a reported total insertion clearance of approximately 1.5~mm. Similarly, Duan et al.\ studied a cable-in-duct installation task using a zero-shot Sim2Real visual–tactile reinforcement learning framework, evaluating robustness by varying pipe diameters (16--25~mm) for fitting a 4~mm fish tape~\cite{duan2025}.

Among learning-based studies in construction, Apolinarska et al.\ (mentioned in Section~\ref{crm_construction}) is one of the few studies deploying reinforcement learning on a large-scale industrial manipulator, evaluating millimeter-scale (1 and 2~mm) geometric clearance for timber lap joint assembly~\cite {apolinarska2021}. Compared to these prior works, the present study investigates sensory--motor policy learning on construction-scale industrial manipulators for timber joinery assembly under substantially tighter sub-millimeter clearances and positional perturbations that exceed these clearances, thereby evaluating the limits of policy robustness in a structurally constrained, contact-rich manipulation task.

\begin{figure*}[!t]
  \centering
   \includegraphics[width=\linewidth]{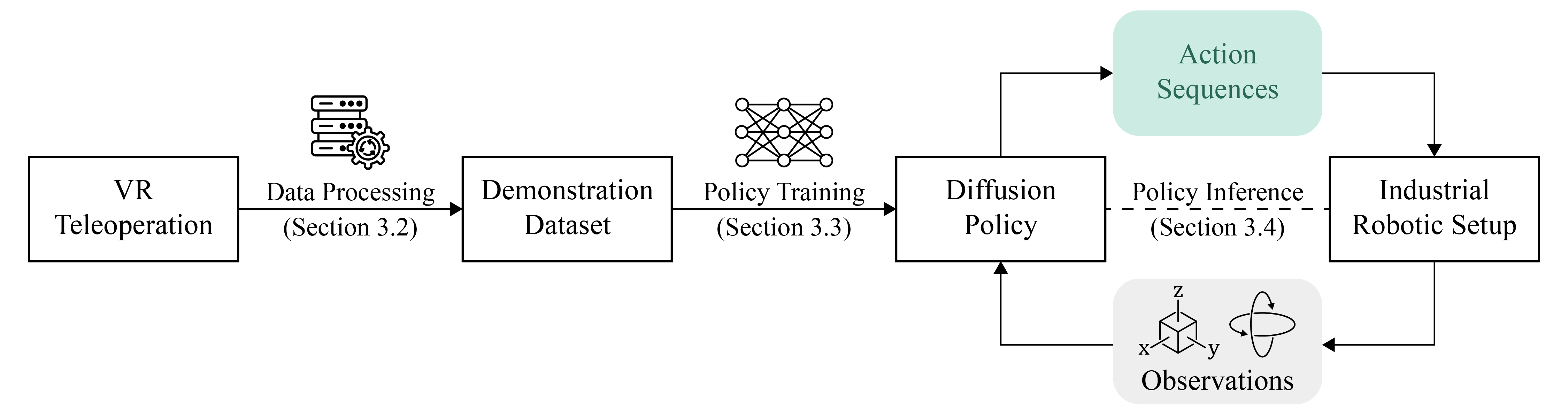}
   \caption{Workflow integrating VR-based teleoperation to collect data,  train diffusion policies, and evaluate learned policies on an industrial robotic setup.}
   \label{fig:methods_overview}
\end{figure*}

%% file: 3_methods.tex
\section{Methods}\label{methods}

This section outlines the experimental setup, including the multi-robot workcell, communication stack, and teleoperation system to collect expert demonstrations using a virtual reality (VR) controller (Section~\ref{setup}), data synchronization and processing to prepare the dataset for training (Section~\ref{data_processing}), a detailed overview of the sensory--motor diffusion policy learning method (Section~\ref{policy_learning}), and the policy inference details to evaluate the trained policies on our experimental setup (Section~\ref{policy_inference}). An overview of the workflow is visualized in Fig.~\ref{fig:methods_overview}.

\begin{figure*}[!t]
  \centering
   \includegraphics[width=\linewidth]{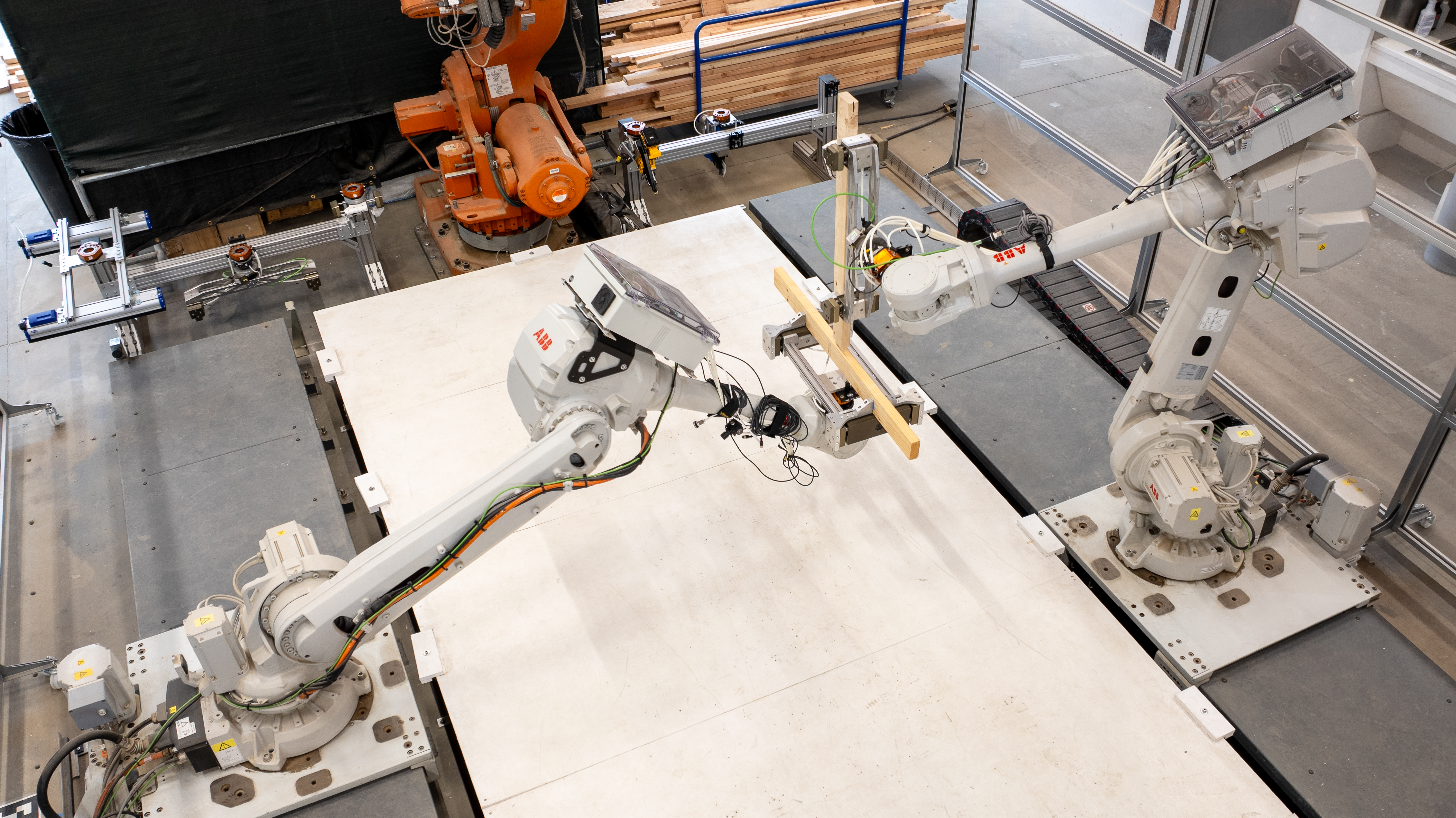}
   \caption{Multi-robot setup consisting of two 6-axis industrial robotic arms.}
   \label{fig:setup}
\end{figure*}

\begin{figure}[!t]
  \centering
   \includegraphics[width=\linewidth]{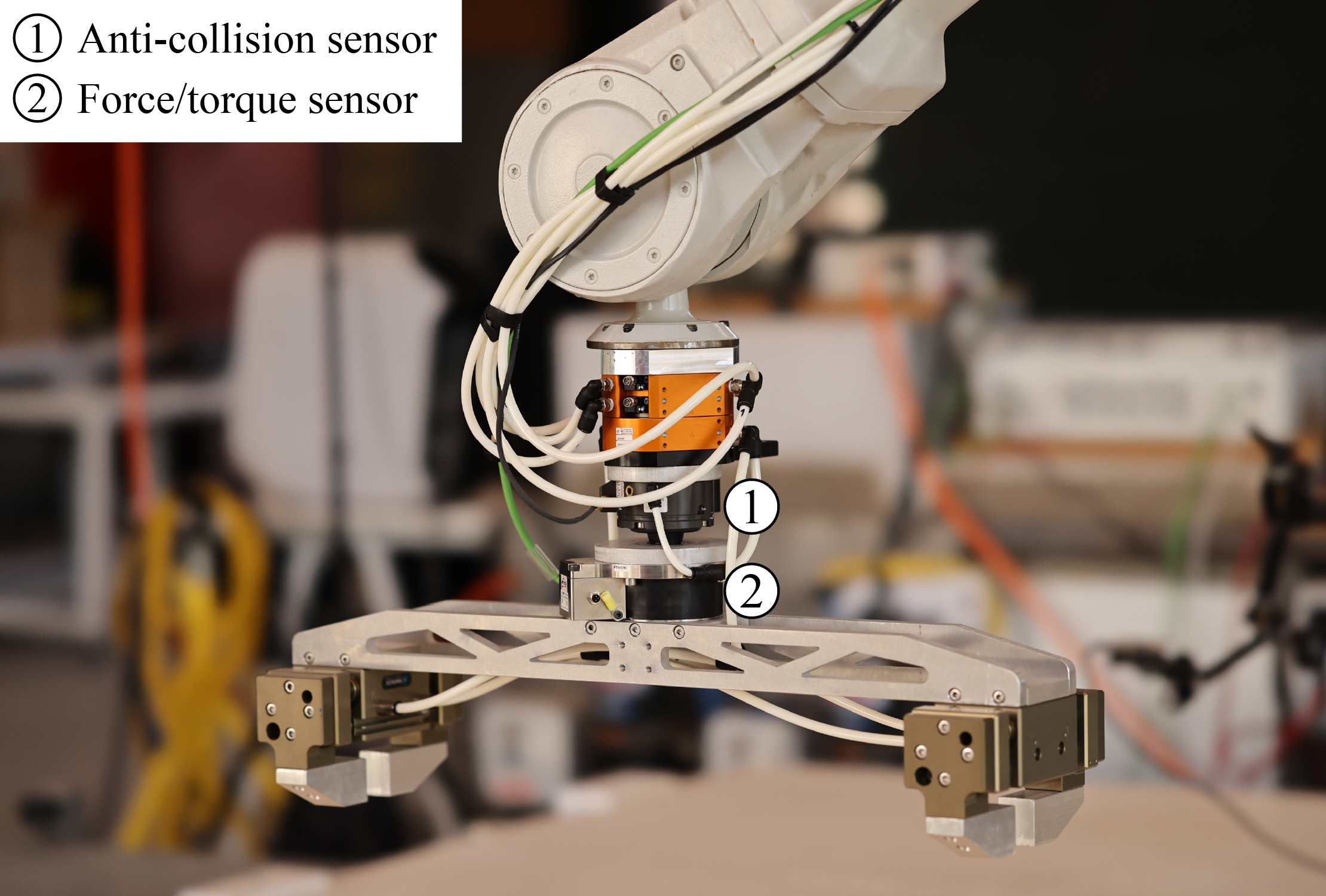}
   \caption{Custom gripper end effector equipped with anti-collision and F/T sensors.}
   \label{fig:gripper}
\end{figure}

\subsection{Experimental setup}\label{setup}

Our experimental setup consists of two primary systems for conducting this research. The first is a multi-robot setup equipped with customized end effectors for contact-rich manipulation and timber assembly. The second is a teleoperation system for manually operating the robotic setup and collecting data during demonstrations. 

The multi-robot setup consists of two six-axis industrial robotic arms\footnote{ABB IRB 4600~\cite{abb}}, each with a 40~kg payload and a 2.55~m reach mounted on linear tracks (Fig.~\ref{fig:setup}). Both robotic arms utilize custom pneumatically-driven end effectors designed for gripping long elements with varying cross-sectional profiles. Because the industrial robotic arms do not provide native access to joint torque values, one of the end effectors includes a six-axis F/T sensor\footnote{ATI Delta IP60, with SI-330-30 calibration~\cite{ati}} to enable F/T data streaming (Fig.~\ref{fig:gripper}). For safety, the same end effector is also equipped with a pneumatic anti-collision sensor\footnote{Schunk OPR 081-P00~\cite{schunk-ac}}, which provides passive mechanical compliance and triggers a protective stop on the robotic arm under excessive force, safeguarding the robotic setup and F/T sensor.

Motion control and pose feedback for the robotic arms utilize ABB's Externally Guided Motion (EGM) feature~\cite{rapid2017}, a UDP-based real-time control interface. F/T data is transferred over EtherCAT to a central Programmable Logic Controller (PLC)\footnote{Beckhoff CX2062~\cite{beckoff}}. The commanded end effector poses are transmitted via EGM to the robot’s low-level controller, which performs inverse kinematics and executes the corresponding joint motions. EGM acts only as the communication interface for Cartesian commands, while joint-level actuation is handled entirely onboard. All data communications are centralized on a desktop computer connected to the central PLC via a local area network (LAN) employing ROS\footnote{ROS 2 Jazzy Jalisco~\cite{ros2}} interfaces and Python wrappers.

\begin{figure}[!ht]
  \centering
   \includegraphics[width=0.8\linewidth]{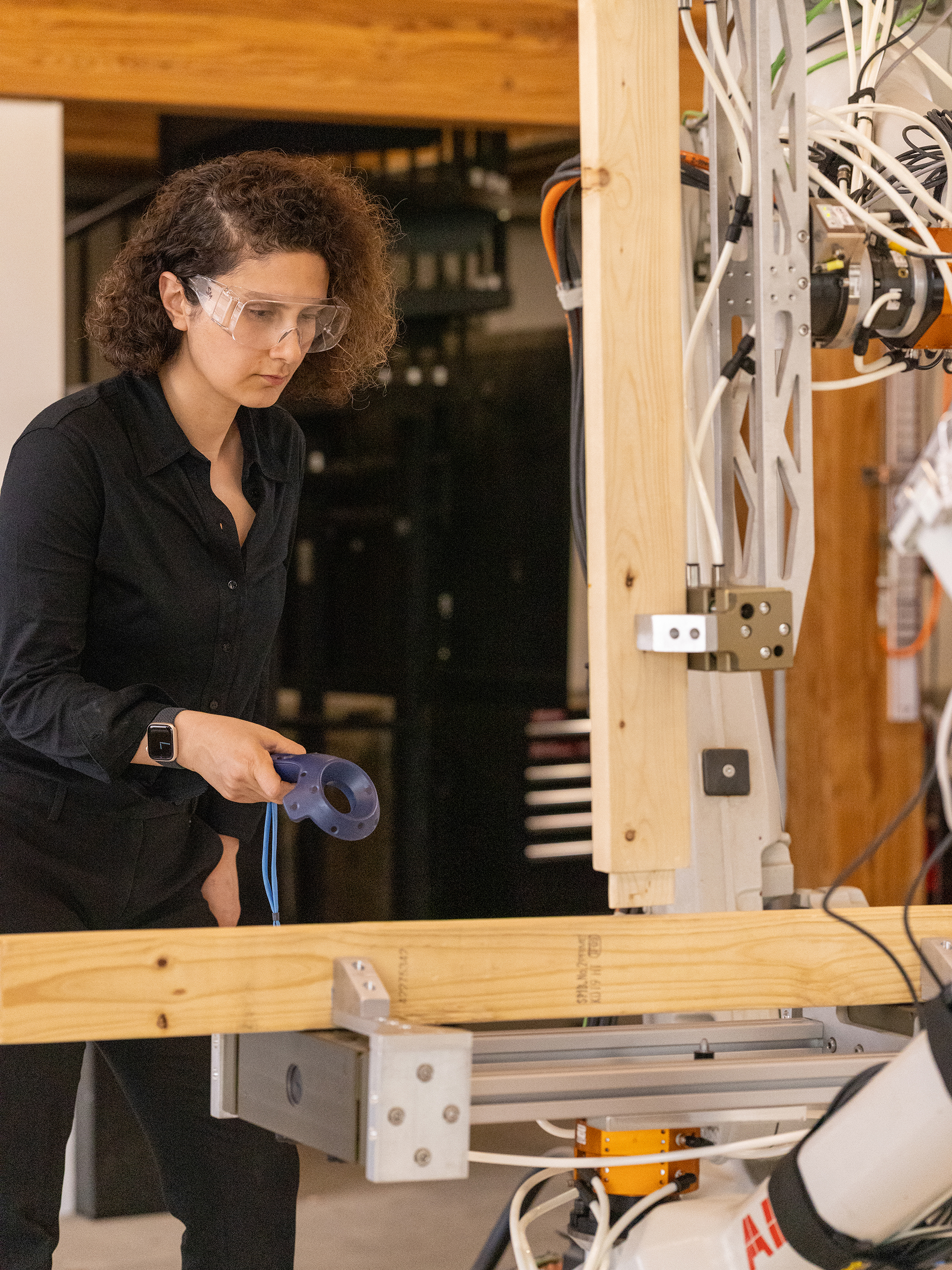}
   \caption{VR-based teleoperation system for collecting human expert demonstrations.}
   \label{fig:demonstration}
\end{figure}

The teleoperation system (Fig.~\ref{fig:demonstration}) implemented in this study is based on a VR interface and follows the framework introduced in our prior work~\cite{ruan2026}. In this system, a six-degree-of-freedom pose is streamed from the VR hand controller\footnote{HTC VIVE Pro 2~\cite{htc}} to the desktop computer via the OpenVR~\cite{openvr} application programming interface (API) and mapped in real time to robot end effector pose commands streamed over EGM. Specifically, we implement a unilateral pose-to-pose transformation that aligns the VR controller frame with the robot tool center point (TCP) frame (details in~\cite{ruan2026}). This transformation ensures consistent directional behavior between human input and robot motion (e.g., forward controller motion always results in motion away from the operator, regardless of physical orientation). 

When the operator presses the hand controller trigger, its current pose and the robot TCP pose are recorded as reference frames. While the trigger remains engaged, subsequent controller motions are interpreted as relative displacements with respect to these references and converted into incremental robot motion commands. Translational and rotational components are processed separately and may be independently scaled, enabling fine-grained control during high-precision, contact-rich operations such as tight-tolerance insertion~\cite{ruan2026}. Additionally, the operator receives haptic feedback through controller vibrations when the measured force from the F/T sensor exceeds a predefined threshold (less than the sensing range maximum value). This cue provides awareness of contact intensity during teleoperation, supporting safer interaction and more controlled force application.

\subsection{Data processing}\label{data_processing}

\begin{figure*}[!h]
  \centering
   \includegraphics[width=1.0\linewidth]{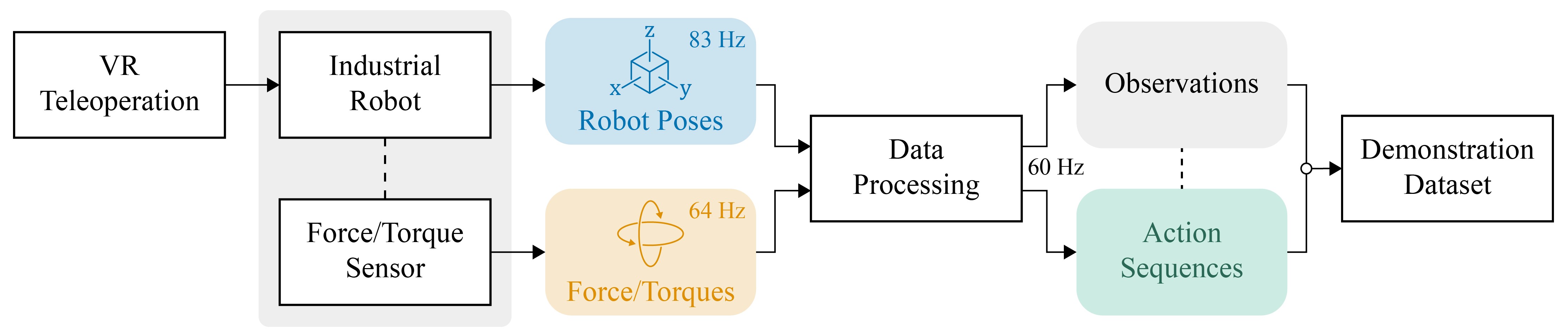}
   \caption{Data collection via VR-based teleoperation.}
   \label{fig:workflow}
\end{figure*}

During teleoperation, the raw data for each demonstration is recorded for each sensor interface (Fig.~\ref{fig:workflow}): end effector TCP pose data is collected from the robot controller as a seven-dimensional (7D) vector (3D position in millimeters + 4D quaternion rotation of the tenon gripper end effector) every 12~ms (approximately 83~Hz), while F/T data is collected as a 6D vector (3D force in newtons + 3D torque in newton-meters) at 64~Hz after passing through an infinite impulse response (IIR) low-pass filter to attenuate frequency components above 64~Hz. This filtering step mitigates aliasing and improves signal quality by removing sensor noise and dynamic vibrations introduced by high-speed motion or structural resonances. To temporally align the different sample rates to a common time grid, the pose and F/T data are interpolated to 60~Hz; position and F/T data are linearly interpolated, while rotation data uses spherical linear interpolation to preserve smooth interpolation.

\begin{figure*}[!h]
  \centering
   \includegraphics[width=1\linewidth]{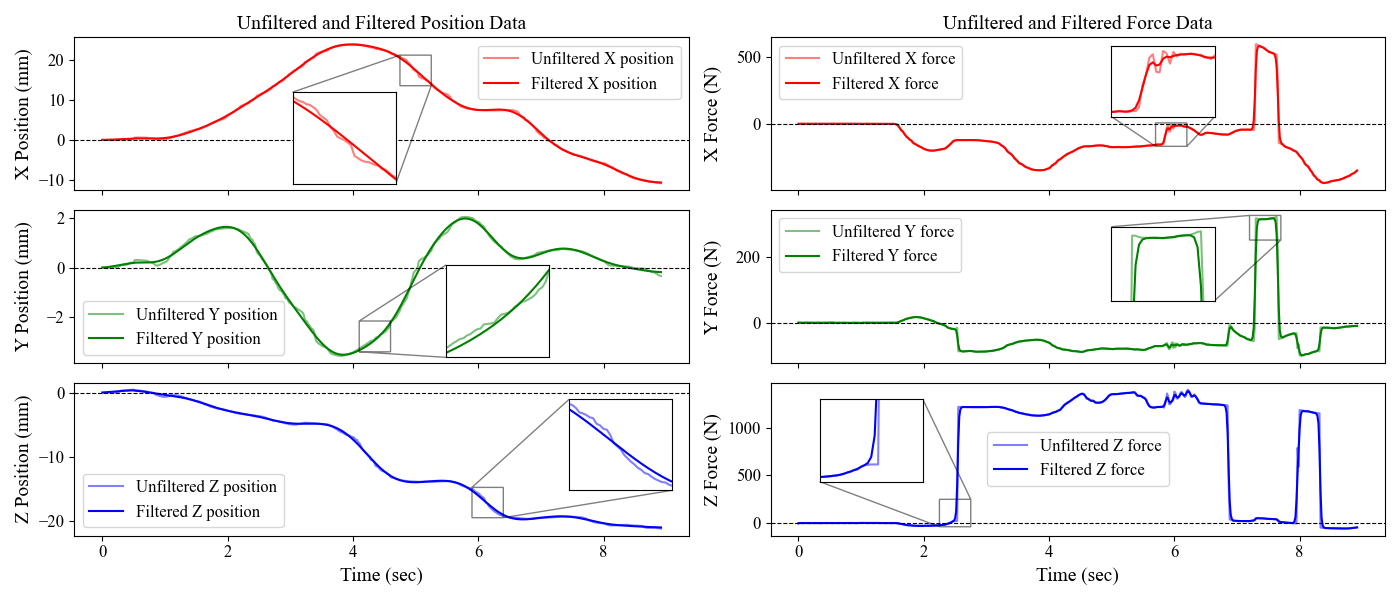}
   \caption{Data filtering using a low-pass Butterworth filter. The visualized trajectory is for an error recovery scenario, demonstrating insertion after a collision with the mortise surface. Only the first 3 dimensions for each pose and F/T data are shown.}
   \label{fig:filtering}
\end{figure*}

Finally, each episode is post-processed to remove periods of inactivity or unintentional stillness and reduce unwanted noise in the demonstrations. Removing idle or near-idle periods is critical as it helps the policy training to focus on active, meaningful interactions rather than noise or moments where no significant movement occurs. We filter the pose and F/T data using a low-pass Butterworth filter~\cite{Butterworth1930}, with filter parameters selected to reduce unwanted high-frequency noise while preserving the underlying movement patterns (Fig.~\ref{fig:filtering}). For the pose data, we used a 4th-order low-pass Butterworth filter with a 1~Hz cutoff, which provides a steeper roll-off and ensures that only low-frequency components characteristic of slower, continuous motion trajectories remain. In contrast, we used a 1st-order low-pass Butterworth filter with a higher 10~Hz cutoff to preserve sharper transitions (e.g., at the moment of contact) while effectively removing high-frequency noise that may come from mechanical oscillations or sensor inaccuracies.

\subsection{Policy training}\label{policy_learning}

We train a CNN-based sensory--motor diffusion policy via action diffusion~\cite{chi2024a}; this means that the policy uses a convolutional backbone for the noise prediction network (as opposed to a transformer-based architecture). The policy training uses a standard DDPM-style training~\cite{ho2020}. During inference, we adopt a DDIM-style sampling procedure~\cite{song2022}, which enables accelerated trajectory generation using fewer denoising steps for efficient action prediction during robot execution (see Section~\ref{policy_inference}). The policy predicts a sequence of robot actions (i.e., the robot TCP pose) conditioned on an observation horizon of robot pose and F/T data. The foundation for our diffusion policy implementation is built upon the work of Bogert et al.~\cite{bogert2024}.

When loading the dataset for training the diffusion policies, trajectories of 64 points are subsampled from each demonstration. This subsampling normalizes the trajectory length for each demonstration, as well as enables faster training iterations. We also convert pose quaternions to a continuous 6D rotation representation (by first converting to rotation matrices, then concatenating the first two columns) to improve training stability~\cite{zhou2019, chi2024a}, resulting in 9D pose vectors. We then apply min-max normalization to scale all features to the range [-1, 1]. Below, we formalize the diffusion policy training on the subsampled and normalized dataset.

Given a pose observation horizon $T_o^p$, F/T observation horizon $T_o^f$, and action prediction horizon $T_p$, each training sample at timestep $t$ consists of an observation feature $\mathbf{O}_t$ concatenating the last $T_o^p$ and $T_o^f$ and a corresponding action sequence $\mathbf{A}_t$ consisting of the next $T_p$ absolute poses. During inference, $\mathbf{A}_t$ is predicted conditioned on $\mathbf{O}_t$. The reverse denoising process for action prediction starts from ${A}^K$ sampled from Gaussian noise. After iterating $K$ denoising steps, we output a desired noise-free action ${A}^0$ at step $0$. The action at each reverse step $k-1$ is calculated as~\cite{chi2024a}:

\begin{equation}
\mathbf{A}_t^{k-1} = \alpha \left[\mathbf{A}_t^k - \gamma \epsilon_\theta + \mathcal{N}\left(0, \sigma^2 I\right)\right]
\end{equation}

\noindent where $\epsilon_\theta$ is the learned predicted noise and $\mathcal{N}(0, \sigma^2 I)$ is Gaussian noise. The $\alpha$, $\gamma$, and $\sigma$ are functions of the iteration step $k$ and are defined through a selected noise schedule. In our case, a squared cosine noise schedule is used~\cite{nichol2021}. The noise prediction network $\epsilon_\theta$ is learned using a 1D CNN U-Net architecture by minimizing the following mean squared error (MSE) loss~\cite{chi2024a}:

\begin{equation}\label{eq:mse_loss}
\mathcal{L} = \sum \left\lVert \epsilon^k - \epsilon_\theta\left(\mathbf{O}_{t},\mathbf{A}^k_t,k\right) \right\rVert
\end{equation}

\noindent where $\epsilon^k$ is the sampled random noise with known variance at iteration $k$. The predicted noise $\epsilon_\theta(\mathbf{O}_{t},\mathbf{A}^k_t,k)$ is a function of $\mathbf{O}_{t}$, the pose and F/T observation feedback preceding time $t$, $\mathbf{A}^k_t$ = $\mathbf{A}^0 + \epsilon^k$ with $\mathbf{A}^0$ sampled from starting noise-free actions, and iteration step $k$. As implemented in~\cite{chi2024a}, the sampled random noise $\epsilon^k$ is added to the noise-free actions $\mathbf{A}_0$ in $K$ forward steps (original DDPM method).

Each policy was trained for a fixed amount of time (1~hour), with each training using a batch size of 128 and lasting at least 1,000~epochs. Only the model weights corresponding to the epoch with the lowest validation loss were used to evaluate performance during rollouts.

\subsection{Policy inference}\label{policy_inference}

To evaluate the performance of our trained policies, we perform rollouts on the experimental setup (Section~\ref{setup}). During policy inference, we specify additional parameters for the number of inference steps $K_{\text{inf}}$, DDIM/DDPM interpolation $\eta$~\cite{song2022}, and action execution steps $T_a$. The number of inference steps ($K_{\text{inf}} \leq K$) controls the number of steps in the reverse denoising process during policy inference. Decreasing $K_{\text{inf}}$ reduces computation and increases rollout speed, which is important for real-time policy execution. However, smaller $K_{\text{inf}}$ may also reduce the quality of noise sampling or action precision. In this work, we adopt DDIM-style sampling (i.e., reduced $K_{\text{inf}}$ with $\eta < 1$) to enable faster action generation during closed-loop execution. This is particularly important for our study, where the policy must repeatedly replan in response to evolving contact interactions and positional misalignment.

The amount of stochasticity during the reverse denoising process is controlled by the interpolation parameter $\eta \in [0, 1]$, where $\eta=1$ is the original DDPM process. This means that when $\eta=0$, the standard deviation of the noise added during sampling is 0, making the reverse process fully deterministic and producing high-fidelity imitations of demonstrations. When $\eta=1$, the reverse process is fully stochastic, introducing broader behavioral coverage through exploration and variability. While DDPM-style sampling ($\eta=1$) provides higher stochasticity, it incurs significantly higher computational cost due to the larger number of denoising steps required during inference.

The number of action execution steps $T_a \leq T_p$ denotes that of the $T_p$ predicted actions at each time step; we execute the first $T_a$ actions. Decreasing the value of $T_a$ increases the frequency of replanning over the receding horizon, thereby increasing the system's responsiveness; however, it may also reduce the speed of rollouts by increasing inference time. For values of $T_a > 2$, we apply average pooling with kernel size $T_a - 1$ across each action dimension, thereby smoothing the action sequence. In this study, we set $T_a = T_p/2$ for all experiments.

In contrast to data collection during demonstrations, observations during inference are not temporally aligned, and instead, we use the latest available pose and F/T data when the observation is retrieved. These observations are normalized using the same normalization parameters as the training dataset before being passed to the policy for inference. The output action sequence is similarly denormalized back into the original scale using the same action normalization parameters during training.

%% file: 4_experiments.tex
\begin{figure*}[h]
  \centering
   \includegraphics[width=1\linewidth]{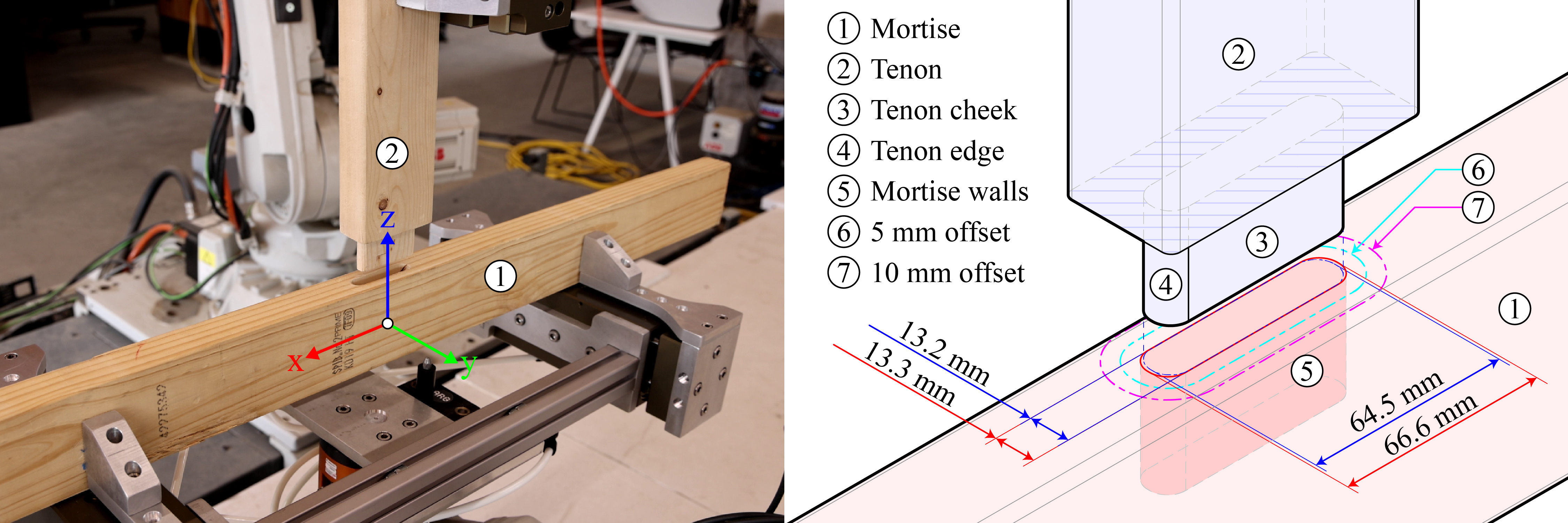}
   \caption{Mortise and tenon assembly task and reference frame (left); terminology and dimensions (right).}
   \label{fig:joint_dims}
\end{figure*}

\section{Experiments}\label{experiments}

This section describes the experimental methodology and evaluation protocol used in this study. We first introduce the mortise and tenon assembly task and joint dimensions (Section~\ref{task}). We then present the data collection procedure used to construct the demonstration datasets (Section~\ref{data_collection}), and the criteria used to measure policy performance (Section~\ref{policy_evaluation}). Next, we outline two experimental phases: Phase~1 focuses on baseline evaluation and parameter initialization under fixed conditions (Section~\ref{phase1_experiments}), while Phase~2 fine-tunes the parameters, examines policy robustness under randomized mortise position perturbations, including additional force ablation and demonstration count studies (Section~\ref{phase2_experiments}). We finally provide a statistical analysis strategy to evaluate the observed differences in policy success rates (Section~\ref{stats_analysis}).

\subsection{Task description} \label{task}

The experimental task is the contact-rich robotic assembly of a vertical mortise and tenon joint as illustrated in Fig.~\ref{fig:joint_dims}. Each robot initially grasps a timber element, and we assume that a motion planner is responsible for placing the grasped tenon at a pre-insertion pose. This pre-insertion pose is constant for all demonstrations and rollouts, located near the mortise, positioned approximately 15~mm above the mortise hole, and subject to an initial angular misalignment of approximately 6~degrees about the y-axis (Fig.~\ref{fig:joint_dims}). Uncertainties associated with grasping are outside the scope of this study and are not considered. The task objective is to fully insert the tenon into the mortise using compliant, contact-aware motion. Only the robot with the grasped tenon is controlled while the mortise pose remains fixed throughout the task. 

The geometric dimensions of the mortise and tenon used in all experiments are shown in Fig.~\ref{fig:joint_dims}. The clearance between the tenon cheeks and the mortise walls is 0.1~mm, representing a tight structural fit typical of traditional timber joinery, while the clearance between the tenon edges and the mortise walls is approximately 2~mm. The tight tolerance of the tenon cheeks governs load transfer and frictional contact during insertion and is the primary source of contact sensitivity in the task. Figure~\ref{fig:joint_dims} also visualizes the boundaries corresponding to the 5~mm and 10~mm offsets used in the Phase 2 experiments (Section~\ref{phase2_experiments}). These offsets mimic realistic deviations arising from positioning errors, tolerance accumulation, and material imperfections commonly encountered in construction-scale robotic assembly~\cite{adel2024, gandia2022}.

\subsection{Data collection}\label{data_collection}

To train the diffusion policies, we created two datasets consisting of expert demonstrations. During data collection, a human demonstrator utilized the teleoperation system (Section~\ref{setup}) to complete the vertical mortise and tenon assembly task. The first dataset consists of 100 demonstrations with fixed starting positions for the mortise and tenon. This dataset is used by the Phase~1 experiments to initialize parameter tuning (Section~\ref{phase1_experiments}).

The second dataset consists of 400 demonstrations with a fixed tenon starting position but a randomized mortise position, simulating fabrication uncertainty. The mortise position was perturbed by applying a planar translational offset sampled in polar coordinates, with direction $\theta \sim U(0, 2\pi)$ and magnitude $r \sim U(0, 10\text{~mm})$, and converted to Cartesian offsets $\left( \Delta x, \Delta y \right) = (r \cos{\theta}, r \sin{\theta})$. This sampling scheme was selected to approximate assembly tolerances, placing greater probability mass near small offsets while still demonstrating larger deviations and boundary cases, and remaining simple to implement directly on the industrial robot controller. To improve the robustness of the learned policy under fabrication uncertainties, we dedicated 100 (25\%) of the demonstrations to error recovery scenarios~\cite{black2026}. In these cases, the demonstrator intentionally initiates contact with the edges of the mortise and slides along the surface to realign and insert the tenon, mimicking realistic recovery behaviors in the presence of misalignment. This dataset is used in the Phase~2 experiments (Section~\ref{phase2_experiments}).

\subsection{Policy evaluation}\label{policy_evaluation}

For each experimental configuration, the policy is trained 4 times with randomized seeds, and policy performance is determined by rolling out 5 times for each trained model, to account for stochasticity during test time, yielding an average success rate (Avg. SR) across 20 combined rollouts. A rollout is recorded as a success if the tenon element fully inserts into the mortise hole without triggering any collision errors (i.e., excessive force at either the collision sensor or the robot joints). A rollout is recorded as a failure if the anti-collision sensor triggers a protective stop due to excessive contact forces exceeding the allowable threshold of the F/T sensor, or if the insertion remains incomplete after approximately 20--30 inference iterations, as determined by the operator. (see examples of successful and failed rollouts in Fig.~\ref{fig:rollouts}).\\

\subsection{Phase~1: Static experiments}\label{phase1_experiments}

The objective of the Phase~1 experiments is to evaluate the applicability and baseline performance of diffusion policies for contact-rich manipulation of the mortise and tenon joint. The policies were trained on the first dataset of 100 demonstrations (Section~\ref{data_collection}). Through repeated experimentation, we identified the best-performing parameter set (reported in Section~\ref{results:phase1}) and used it to initialize parameter tuning in the Phase~2 experiments described in the following subsection.

\subsection{Phase~2: Uncertainty experiments}\label{phase2_experiments}

The objective of the Phase~2 experiments is to evaluate the robustness of diffusion policies in handling uncertainties arising from positioning errors, tolerance accumulation, and material imperfections commonly encountered in construction-scale assembly~\cite{adel2024, gandia2022}. Specifically, we assess policy performance when the mortise is subjected to randomized positional offsets to simulate fabrication uncertainty. The mortise orientation is not randomized and remains constant. The policies for these experiments were trained using the second dataset of 400 demonstrations (Section~\ref{data_collection}).

During rollouts, we evaluated each policy with varying mortise position perturbations: 0~mm (no perturbation), 5~mm, and 10~mm. These values were selected to evaluate the policies under no uncertainty (0~mm), uncertainty within the demonstration distribution (5~mm), and uncertainty at the edge of the demonstration distribution (10~mm). For each offset condition, the mortise position was uniformly randomized along the circumference of a circle with a radius equal to the specified distance. The Avg. SR of a policy was computed separately for each offset distance, and its overall performance was reported as the average total success rate (Avg. Total SR) across all three conditions.

In the first part of the Phase~2 experiments, we systematically tuned the diffusion policy parameters to maximize the Avg. Total SR. Starting from a candidate configuration derived from the Phase~1 experiments (Section~\ref{phase1_experiments}), we performed sequential, greedy parameter tuning, optimizing one parameter at a time while holding others fixed. The resulting best-performing policy, with the highest Avg. Total SR is referred to as the \textit{full model}. The impact of individual parameters on policy performance is reported in Section~\ref{results:phase2:full}.

In the second part of the Phase~2 experiments, we conducted two studies to assess the impact of F/T observations and demonstration count on policy performance. In the F/T ablation study, we evaluated the performance of the full model with F/T inputs masked (i.e., F/T values set to 0) during inference, as well as a variant trained only on pose observations. For the demonstration count, we trained policies using 25, 50, 100, and 200 demonstrations (each including 25\% error recovery cases as explained in Section~\ref{data_collection}), while keeping all other parameters identical to the full model. The results of these studies are presented in Sections~\ref{results:phase2:ft-ablations}~and~\ref{results:phase2:demo-count}.

\subsection{Statistical analysis}\label{stats_analysis}

To assess whether observed differences in success rates are statistically significant, we model rollout outcomes using a binomial generalized linear model (GLM) with a logit link function~\cite{mccullagh1989}. This formulation is appropriate for binary success/failure outcomes and allows us to compare policies while accounting for the number of rollouts per configuration.

We first test whether any differences exist across policies using a likelihood-ratio test comparing a model with policy as a categorical factor to a null (intercept-only) model. When this test indicates potential differences, we perform pairwise Wald tests on log-odds differences, with Holm--Bonferroni correction applied to control for multiple comparisons~\cite{holm1979}.

For the Phase~2 demonstration count study, we instead treat the number of demonstrations as a continuous variable by fitting a binomial GLM with log$_2$-scaled demonstration count. This allows us to test whether increasing the number of demonstrations leads to a consistent improvement in success rate.

%% file: 5_results.tex
\section{Results and discussion}\label{results}

\begin{figure}[h]
  \centering
   \includegraphics[width=1\linewidth]{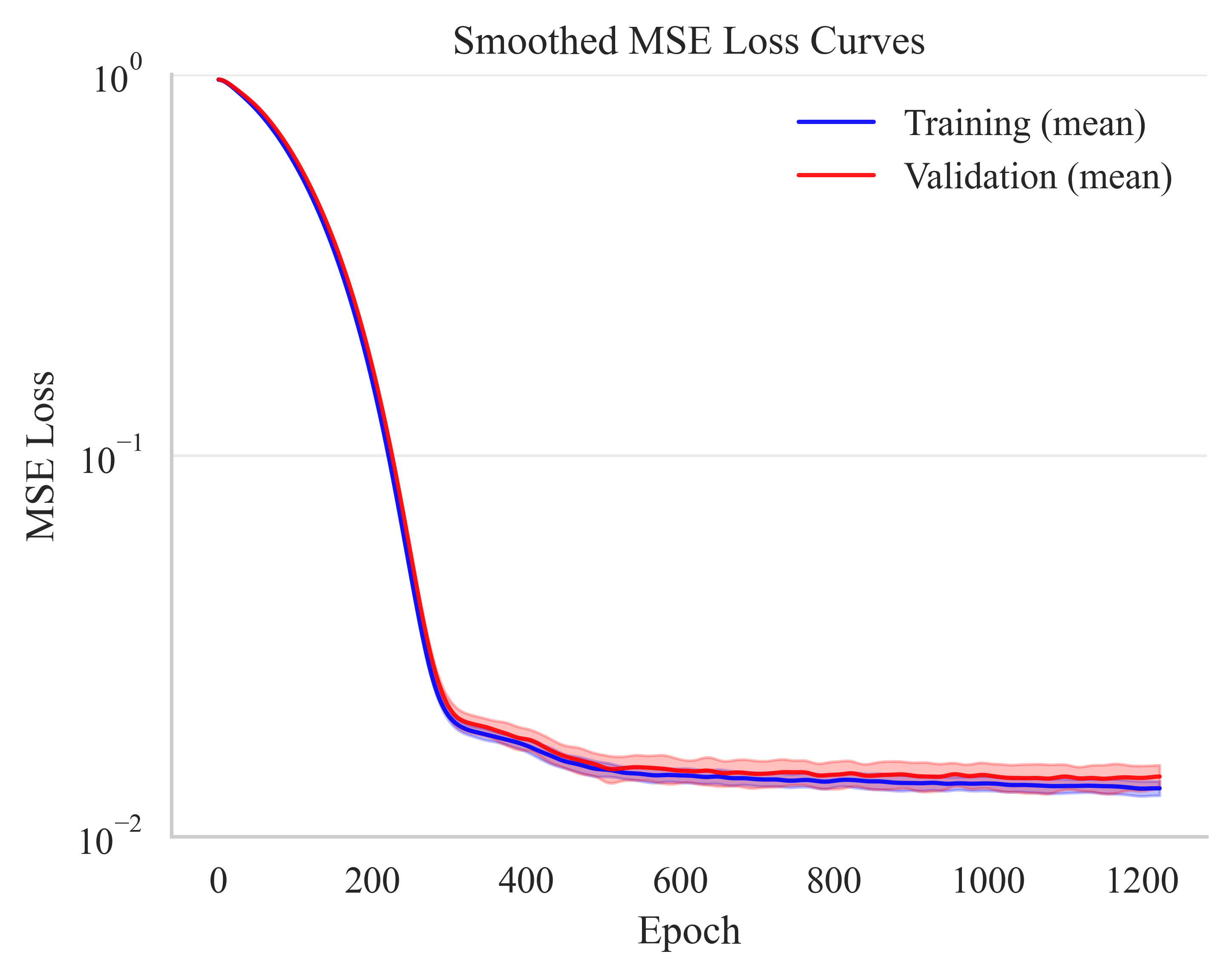}
   \caption{Smoothed mean squared error (MSE) training and validation loss curves for Policy 4, averaged and then smoothed across all training iterations. The shaded area represents $\pm 1$ standard deviation.}
   \label{fig:loss_curves}
\end{figure}

Across all experiments, policy training exhibited consistent and stable convergence. Training and validation MSE losses remained closely aligned throughout policy training, suggesting minimal overfitting. Both losses decreased steadily over the first 300 epochs, followed by a gradual plateau beginning at approximately 600 epochs. An example of the MSE loss curves is illustrated in Fig.~\ref{fig:loss_curves} for the full model (see Section \ref{results:phase2:full}). The loss curves were first averaged across all 4 training iterations, then smoothed using a Gaussian filter with a standard deviation $\sigma=10$ for visualization.

\begin{figure*}[t]
  \centering
   \includegraphics[width=1\linewidth]{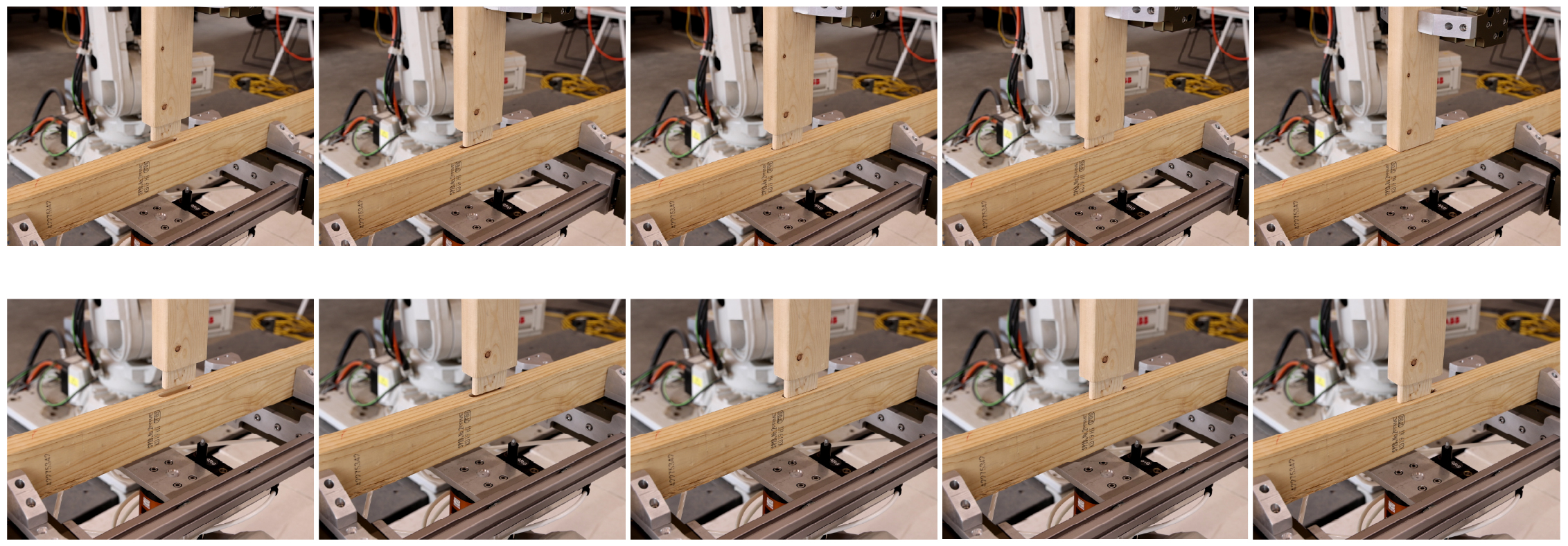}
   \caption{Example of an insertion sequence for a successful rollout (top row) and an unsuccessful rollout (bottom row).}
   \label{fig:rollouts}
\end{figure*}

Fig.~\ref{fig:rollouts} illustrates two example rollout trajectories: the top sequence shows a successful rollout where the tenon was first reoriented and then smoothly inserted into the mortise; the bottom sequence shows an unsuccessful rollout where the tenon did not finish reorienting before hitting the edge of the mortise and was unable to recover correctly.

\subsection{Phase~1: Static experiments}\label{results:phase1}

In Phase 1, after repeated testing and initial parameter tuning, the policy achieved a 100\% average success rate over 20 rollouts using the parameter values listed in Table~\ref{tab:params}. This performance demonstrates the capability of diffusion policies to reliably assemble a tight-fitting timber joint under no uncertainty. These parameter values were subsequently used as the initialization point for parameter tuning in the Phase~2 experiments.

While these results demonstrate that diffusion policies can reliably perform the assembly task under nominal conditions, they should be interpreted within the scope of the controlled experimental setup. In particular, the task does not capture the broader variability and complexity of real-world construction scenarios, where additional sources of uncertainty, multi-step dependencies, and diverse joint geometries may significantly affect performance.

\begin{table}[!ht]
    \centering
    \caption{Parameters for the Phase~1 experiments.}
    \label{tab:params}
    \begin{tabular}{llr}
        \toprule
        Symbol & Description & Value \\
        \midrule
        $L_r$ & Learning rate & $1\mathrm{e}{-5}$ \\
        $W_d$ & Weight decay rate & $1\mathrm{e}{-6}$ \\
        $T^p_o$ & Pose observation horizon & 1 \\
        $T^f_o$ & Force observation horizon & 1 \\
        $T_p$ & Action prediction horizon & 8 \\
        $T_a$ & Action execution steps & 4 \\
        $K$ & Forward noising steps & 128 \\
        $K_{\text{inf}}$ & Inference/denoising steps & 32 \\
        $\eta$ & DDIM/DDPM interpolation & 0.5 \\
        \bottomrule
    \end{tabular}
\end{table}

The Phase~1 experiments revealed that a practical challenge arose from the gradual physical degradation of the experimental setup as a consequence of the large contact forces repeatedly exerted on the timber mortise and tenon. This degradation manifested in several ways: wear on contact surfaces (increasing tolerance and making insertion easier), small chips breaking off the tenon edge (slightly decreasing tolerance due to glue reattachment), and minor shifts of the mortise and tenon within the gripper grasp. Such occurrences are expected in real-world construction environments, but they also introduce additional sources of uncertainty. Future research should explicitly account for these fabrication-induced variations when developing and evaluating assembly policies.

\begin{table*}[h] 
\centering
\begin{threeparttable}
\caption{Success rates for training parameter tuning.}
\label{tab:training_params_sr}
\begin{tabular}{ccccccccccc}
\toprule
Policy & \# Demos & $T_p$ & $T^p_o$ & $T^f_o$ & $K_{\text{inf}}$ & $\eta$ & $T_a$ & $D$ (mm) & Avg. SR (\%) & Avg. Total SR (\%)\\
\midrule
1 & 400 & 8 & \cellcolor{gray!20} 1 & \cellcolor{gray!20} 1 & 32 & 0.5 & 4 & 0 & 90 & \cellcolor{gray!20} 70\\
& & & & & & & & 5 & 55 \\
& & & & & & & & 10 & 65 \\
\midrule
2 & 400 & 8 & \textbf{2} & 1 & 32 & 0.5 & 4 & 0 & 85 & 63 \\
& & & & & & & & 5 & 65 \\
& & & & & & & & 10 & 40\\
\midrule
3 & 400 & 8 & \textbf{2} & \textbf{2} & 32 & 0.5 & 4 & 0 & 80 & 60\\
& & & & & & & & 5 & 60 \\
& & & & & & & & 10 & 40 \\
\midrule
\midrule
4 & 400 & \cellcolor{yellow!30} \textbf{16} & \cellcolor{yellow!30}1 & \cellcolor{yellow!30}1 & 32 & 0.5 & \textbf{8} & 0 & 100 & \cellcolor{yellow!30} 75\\
& & & & & & & & 5 & 65 \\
& & & & & & & & 10 & 60 \\
\bottomrule
\end{tabular}
\begin{tablenotes}
\footnotesize
\item $D$ = randomized hole offset.
\end{tablenotes}
\end{threeparttable}
\end{table*}

\subsection{Phase~2: Uncertainty experiments}\label{results:phase2}

In Phase 2, we first perform sequential parameter tuning, starting from the Phase 1 configuration to identify an empirically strong diffusion policy for evaluation under positional uncertainty (referred to as the full model). Second, we analyze the impact of F/T observations and the number of demonstrations on policy performance.

\subsubsection{Full model parameter tuning}\label{results:phase2:full}

The sequential parameter tuning during the Phase~2 experiments was limited to three stages: 1)~pose and F/T observation horizons; 2)~action prediction horizon; 3)~number of inference steps and DDIM/DDPM interpolation. The learning rate, weight decay, and forward noising steps were fixed at the values determined during the Phase~1 experiments. This tuning procedure was intended to identify a strong empirical configuration under the practical constraints of real-robot evaluation, rather than to perform an exhaustive parameter search. As a result, possible interactions among parameters were not fully explored; a more comprehensive analysis of parameter interactions remains a direction for future work.

Table~\ref{tab:training_params_sr} summarizes the results from the first two stages. The highest Avg. Total SR of 75\%, computed over the three mortise offset conditions, was achieved using an observation horizon of~1 for both pose and F/T modalities combined with a longer action prediction horizon of~16. This configuration suggests that, for the perturbations tested, the policy benefited from focusing on the most recent state information while planning actions further into the future, potentially allowing it to better anticipate and correct for deviations introduced by positional offsets. However, the results indicate that there is no statistically significant difference between the evaluated parameter configurations at the 95\% confidence level ($p=0.29$), suggesting that the observed variation in success rates may be attributed to stochasticity in training and evaluation rather than systematic performance gains. Although the differences are not statistically significant, we proceed with the configuration achieving the highest average success rate for subsequent experiments, as it provides the best empirical performance under the evaluated conditions.

Table~\ref{tab:inference_params_sr} summarizes the final stage of parameter tuning, evaluating different combinations of inference step $K_{inf}$ and DDIM/DDPM interpolation $\eta$. These tests were conducted only for mortise offsets of 5~mm, as the overall trend in success rates was consistent across other offsets. The initial parameter values selected from the Phase~1 experiments (i.e., $K_{inf}$ = 32 and $\eta$ = 0.5) achieved the highest performance, with a 65\% average success rate, substantially outperforming other tested configurations.

\begin{table*} [h] 
\centering
\begin{threeparttable}
\caption{Success rates for inference parameter tuning.}
\label{tab:inference_params_sr}
\begin{tabular}{ccccccccccccc}
\toprule
Policy & \# Demos & $T_p$ & $T^p_o$ & $T^f_o$ & $T_a$ & $D$ (mm) & $K_{\text{inf}}$ & $\eta$ & Avg. SR (\%)\\
\midrule
4 & 400 & 16 & 1 & 1 & 8 & 5 & \cellcolor{yellow!30}32 & \cellcolor{yellow!30}0.5 & \cellcolor{yellow!30}65 \\
& & & & & & & \textbf{16} & 0.5 & 35 \\
& & & & & & & \textbf{64} & 0.5 & 40 \\
& & & & & & & 32 & \textbf{1} & 35 \\
& & & & & & & 32 & \textbf{0.25} & 50 \\
\bottomrule
\end{tabular}
\begin{tablenotes}
\footnotesize
\item $D$ = randomized hole offset.
\end{tablenotes}
\end{threeparttable}
\end{table*}

To contextualize these results, we benchmarked policy inference time for representative configurations on the deployment computer. After 10 warm-up iterations, inference time was measured over 100 forward passes. The selected configuration (Policy 4, with $K_{\text{inf}}=32$, $\eta=0.5$) achieved a mean inference time of $408 \pm 10.6$~ms. Reducing the number of denoising steps to $K_{\text{inf}}=16$ further decreased inference time to $297 \pm 8.2$~ms, but at the cost of reduced success rate (35\%). In contrast, full DDPM-style sampling ($K_{\text{inf}}=K=128$, $\eta=1.0$) increased inference time substantially to $1550 \pm 18.8$~ms. These results highlight the tradeoff between computational cost and task performance, and support the use of accelerated DDIM-style sampling as a practical compromise for real-time, contact-rich manipulation.

\subsubsection{Force/torque ablation}\label{results:phase2:ft-ablations}

\begin{figure}[!b]
  \centering
   \includegraphics[width=1\linewidth]{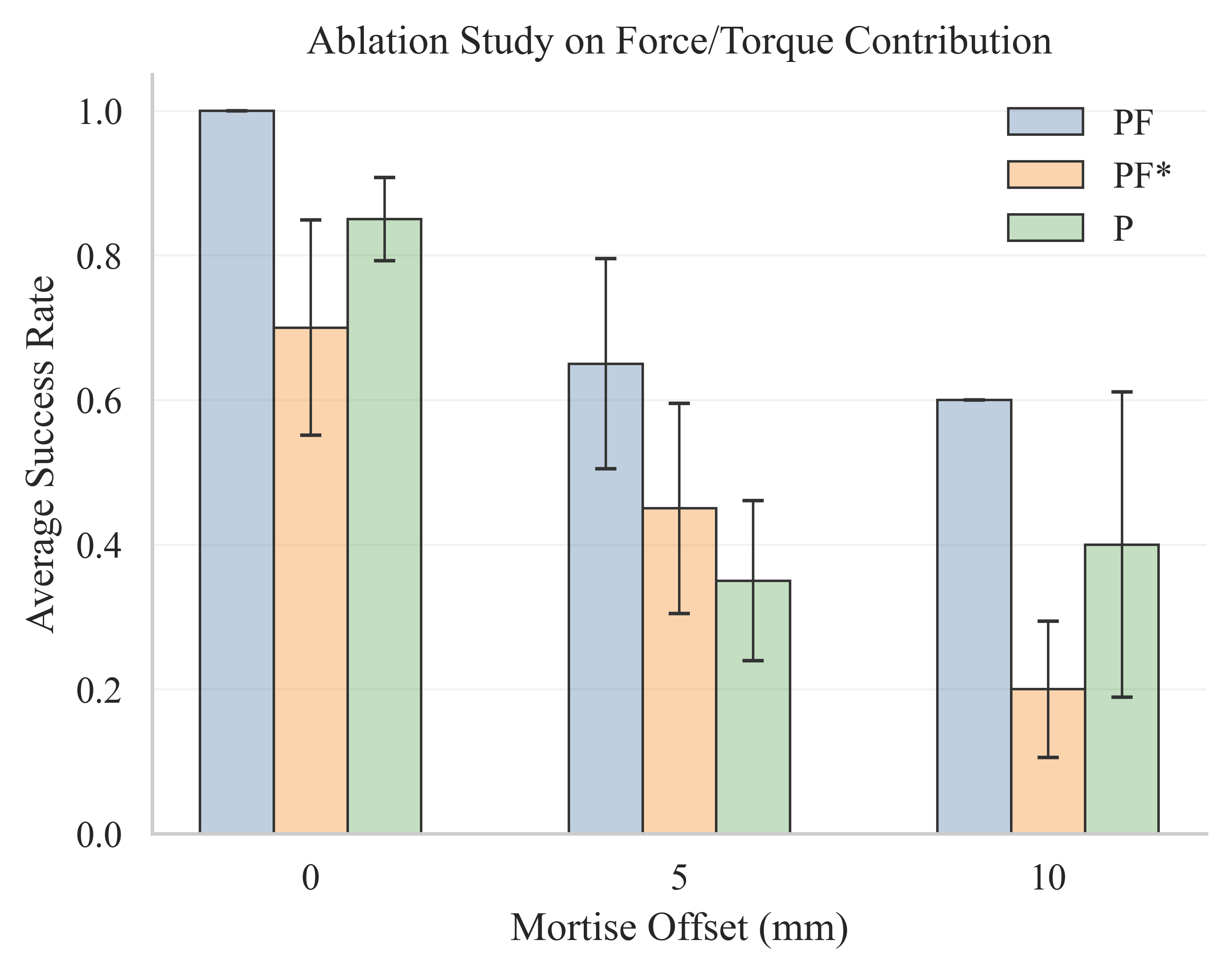}
   \caption{Success rates of diffusion policies with force masked (PF*) and trained only on pose data (P) compared to the full model (PF), evaluated at three mortise offsets (0, 5, and 10~mm). Error bars indicate the standard error of the mean (SEM) computed across the 4 independently trained models for each set of parameters.}
   \label{fig:ft_ablation}
\end{figure}

Table~\ref{tab:ft_ablation_sr} summarizes the results of the F/T ablation study, and Fig.~\ref{fig:ft_ablation} illustrates the contribution of F/T feedback to task performance. At the aggregate level, both ablated policies underperformed relative to the full model. This difference is statistically significant (Section~\ref{stats_analysis}), with the likelihood-ratio test indicating a significant effect of policy choice ($p=0.0023$). Pairwise comparisons further show that the full model significantly outperforms both the masked and pose-only variants (Holm-corrected one-sided tests, $p=0.0010$ and $p=0.0073$, respectively), confirming the critical contribution of force feedback to performance.

The pose-only policy achieved a higher total success rate than the force-masked full model (53\% compared with 45\%). However, this difference was not statistically significant in a direct Holm-corrected two-sided comparison ($p=0.3618$). Thus, the observed difference between the two ablated variants should be interpreted as inconclusive rather than as evidence that pose-only feedback provides a reliable advantage. Qualitatively, the force-masked full model frequently collided with the mortise, suggesting that a policy trained with F/T observations depends on this modality for contact-aware recovery. Overall, these results support the importance of F/T feedback for consistent performance under positional perturbations.

\begin{table*} [h] 
\centering
\begin{threeparttable}
\caption{Success rates for the F/T ablation study.}
\label{tab:ft_ablation_sr}
\begin{tabular}{ccccccccccccc}
\toprule
Policy & Modality & \# Demos & $T_p$ & $T^p_o$ & $T^f_o$ & $K_{\text{inf}}$ & $\eta$ & $T_a$ & D (mm) & Avg. SR (\%) & Avg. Total SR (\%)\\
\midrule
4 & PF & 400 & 16 & 1 & 1 & 32 & 0.5 & 8 & 0 & 100 & 75\\
& & & & & & & & & 5 & 65 \\
& & & & & & & & & 10 & 60 \\
\midrule
4 & PF* & 400 & 16 & 1 & 1 & 32 & 0.5 & 8 & 0 & 70 & 45\\
& & & & & & & & & 5 & 45 \\
& & & & & & & & & 10 & 20\\
\midrule
5 & P & 400 & 16 & 1 & 1 & 32 & 0.5 & 8 & 0 & 85 & 53\\
& & & & & & & & & 5 & 35 \\
& & & & & & & & & 10 & 40 \\
\bottomrule
\end{tabular}
\begin{tablenotes}
\footnotesize
\item $D$ = randomized hole offset; PF = models trained with pose and force data; P = models trained with pose-only data.
\item *The forces were masked during inference.
\end{tablenotes}
\end{threeparttable}
\end{table*}

\begin{figure}[!b]
  \centering
   \includegraphics[width=1\linewidth]{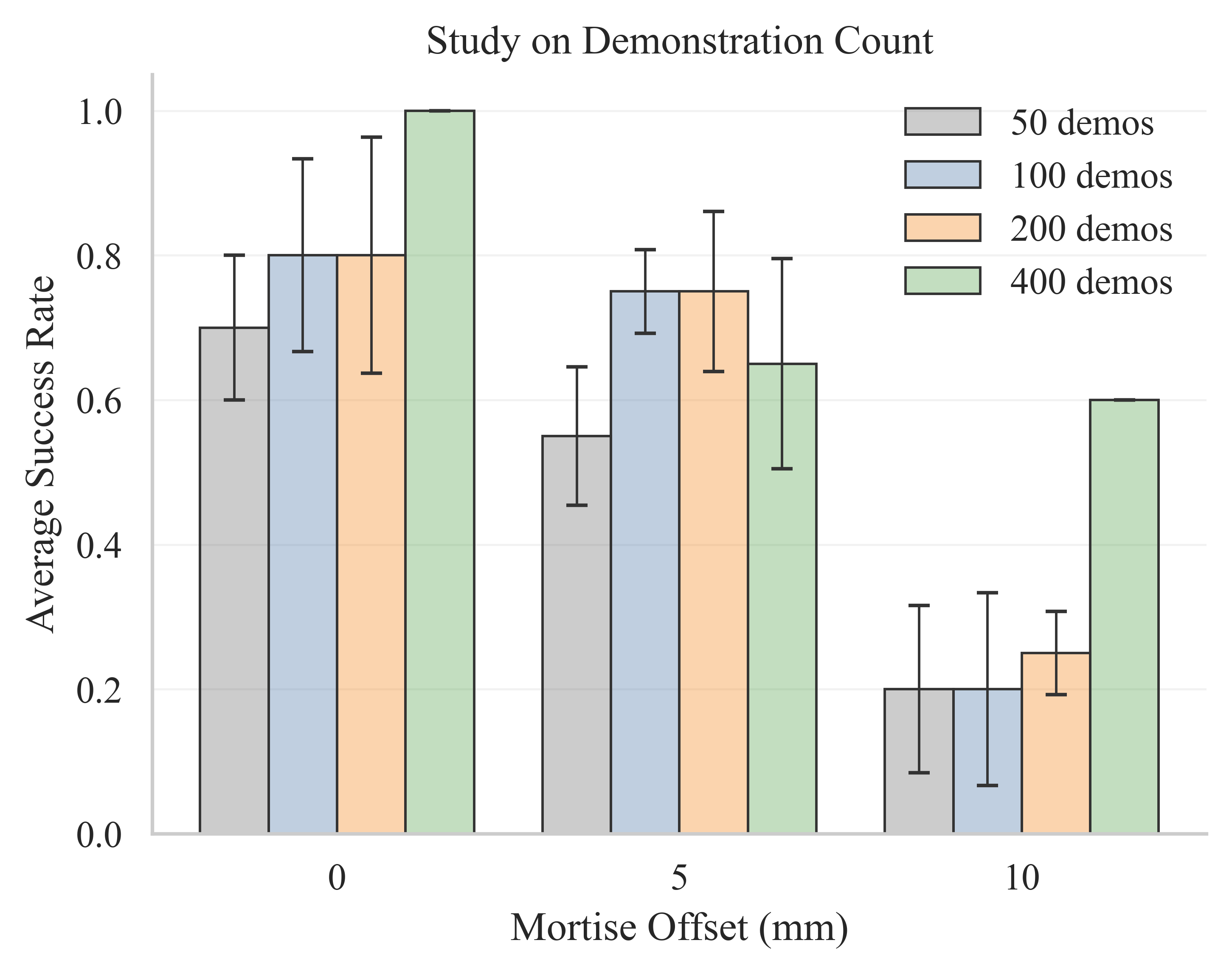}
   \caption{Success rates of diffusion policies trained with different numbers of demonstrations (50, 100, 200, and 400) evaluated at three mortise offsets (0, 5, and 10~mm). Error bars indicate the standard error of the mean (SEM) computed across the 4 independently trained models for each set of parameters.}
   \label{fig:demo_count}
\end{figure}

\subsubsection{Demonstration count}\label{results:phase2:demo-count}

Table~\ref{tab:demo_count_sr} summarizes the results of the demonstration quantity study. Policies trained on only 25 demonstrations failed to fully converge and achieved no success during rollouts, and were therefore excluded from the reported results. Fig.~\ref{fig:demo_count} shows a pronounced drop in success rate at the 10~mm mortise offset when the number of demonstrations was reduced, likely due to insufficient coverage of the state-action space in the demonstration data. Consistent with this observation, we find a statistically significant positive trend between demonstration count and success rate (Section~\ref{stats_analysis}). A binomial GLM with log$_2$-scaled demonstration count yields a significant effect ($p=0.0036$), with an estimated odds ratio of 1.42 per doubling of demonstrations, meaning that each time the number of demonstrations is doubled, the likelihood of successful task completion increases by approximately 42\%.

The results suggest a performance inflection point between 200 and 400 demonstrations, beyond which the policy gains a substantial improvement in its ability to handle positional uncertainty. This trend indicates that under higher uncertainty, diffusion policies may require a critical mass of demonstrations to adequately represent recovery behaviors.

Notably, the total number of demonstrations required to achieve robust performance was much higher than anticipated. This is likely attributable to the trajectory subsampling strategy used during policy training (Section~\ref{policy_learning}), which significantly reduced the total number of training samples available to the policy. Future research should utilize alternative data preprocessing strategies to maintain a sufficient quantity of training data to enable more reliable policy learning.

\begin{table*}[h] 
\centering
\begin{threeparttable}
\caption{Success rates for various demonstration counts.}
\label{tab:demo_count_sr}
\begin{tabular}{ccccccccccc}
\toprule
Policy & \# Demos & $T_p$ & $T^p_o$ & $T^f_o$ & $K_{\text{inf}}$ & $\eta$ & $T_a$ & $D$ (mm) & Avg. SR (\%) & Avg. Total SR (\%)\\
\midrule
4 & 400 & 16 & 1 & 1 & 32
& 0.5 & 8 & 0 & 100 & 75\\
& & & & & & & & 5 & 65 \\
& & & & & & & & 10 & 60 \\
\midrule
6 & \textbf{200} & 16 & 1 & 1 & 32 & 0.5 & 8 & 0 & 80 & 60\\
& & & & & & & & 5 & 75 \\
& & & & & & & & 10 & 25\\
\midrule
7 & \textbf{100} & 16 & 1 & 1 & 32 & 0.5 & 8 & 0 & 80 & 58\\
& & & & & & & & 5 & 75 \\
& & & & & & & & 10 & 20 \\
\midrule
8 & \textbf{50} & 16 & 1 & 1 & 32 & 0.5 & 8 & 0 & 70 & 48\\
& & & & & & & & 5 & 55 \\
& & & & & & & & 10 & 20 \\
\bottomrule
\end{tabular}
\begin{tablenotes}
\footnotesize
\item $D$ = randomized hole offset.
\end{tablenotes}
\end{threeparttable}
\end{table*}

%% file: 6_conclusion.tex
\section{Conclusion}\label{conclusion}

The experimental results presented in this work demonstrate that diffusion policies can achieve stable convergence and high success rates for a contact-rich assembly task under tight tolerances using construction-scale industrial robotic arms. In the Phase~1 experiments, we established a baseline by identifying parameter settings capable of achieving 100\% average success rate in a deterministic mortise and tenon joint assembly task. In the Phase~2 experiments, we evaluated policy robustness under controlled mortise position perturbations, identifying parameter configurations that improved generalization to offsets of the mortise position, quantifying the critical role of force feedback, and determining the demonstration quantity threshold necessary for reliable performance.

These findings validate the applicability of diffusion policies to building-scale robotic assembly using industrial robots and systematically characterize their robustness under fabrication uncertainty. Beyond the performative metrics, the experiments also yielded actionable insights for parameter tuning, sensor fusion, and demonstration count, forming a generalizable methodology for contact-rich construction and assembly tasks.

While the presented methodology suggests potential applicability to related contact-rich assembly tasks, such as other timber joints, pipe fitting, or light metal framing, further validation across a broader range of geometries, materials, and assembly conditions is required to assess its generality. More broadly, robotic timber joinery in construction remains an open challenge, as real-world deployment must address additional complexities, including  combination of twisting and inserting and and integration within multi-step construction workflows. Accordingly, this study should be viewed as an initial step toward understanding how sensory--motor control policies can support contact-rich robotic assembly under positional misalignments. In doing so, this work advances robotic construction under uncertainty while contributing to safer and more efficient building practices, positioning robots as capable collaborators in the evolving construction workforce.

\subsection{Limitations and future work}

While this work demonstrated the feasibility and robustness of applying sensory--motor diffusion policy learning to contact-rich robotic assembly at construction scale, several limitations remain; our approach relies exclusively on robot pose and force/torque data to train the policy and evaluate robustness, whereas diffusion policies for robot planning and control are commonly formulated as visuomotor policies using end-to-end mappings from RGB or tactile images to robot commands~\cite{chi2024a, chi2024b}. Our choice of the sensory input space increases training speed but may limit the model’s ability to generalize to visually complex or occlusion-prone construction scenarios, such as joint assembly in cluttered framing environments or multi-alignment tasks. In particular, the current study focuses on a single joint typology and an orthogonal insertion process, which does not fully capture the contact dynamics encountered in more complex timber joinery configurations. Real-world construction scenarios often involve multi-axis insertions, compound joint geometries, and sequential assembly steps that introduce additional coordination and error propagation challenges.

Also, the sequential parameter tuning and limited rollout sample size limit our ability to detect statistically significant differences between parameter configurations; future work should investigate more systematic tuning strategies that account for parameter interactions and use larger evaluation sample sizes.

In addition, the current policy evaluation focuses on perturbations within the range of the demonstration data, including cases at the boundary of this range (i.e., offsets up to 10 mm along the perimeter of the sampled region); extending to larger, out-of-distribution, and non-planar deviations may further challenge policy robustness, particularly in the absence of visual perception. Incorporating visual sensing could enable the system to handle larger misalignments by improving generalization beyond the local contact regime. Moreover, incorporating expressive state-of-the-art learning models, such as diffusion transformers or vision-language-action models (VLAs)~\cite{black2026, gr2025}, enables better generalization, scalability, and transfer across tasks. These capabilities are particularly relevant for building-scale construction where robots must integrate multiple sensing modalities and operate across varied materials, such as combining timber, metal fasteners, and composite claddings in a single assembly sequence.

Another limitation stems from our reliance on single-task behavior cloning from human demonstrations, which introduces both practical and algorithmic challenges. Collecting high-quality demonstrations and conducting real-world rollouts is time-intensive and can cause material degradation, especially in contact-sensitive scenarios or fragile materials like softwoods, architectural veneers, or insulation panels. Furthermore, the demonstration count study suggests that achieving robust performance under larger perturbations requires a relatively large number of demonstrations. Moreover, learning through behavior cloning is inherently bounded by the quality and diversity of human performance, raising the question of how to scale beyond demonstrator capabilities. This is especially critical for more dexterous and varied AEC tasks, such as assembling shingle patterns on curved surfaces, handling flexible ductwork, or manipulating deformable architectural fabrics~\cite{craney2020, graser2021}. Future work could address these limitations by exploring hybrid or hierarchical learning paradigms that combine demonstration-based training with simulation pretraining or reinforcement learning, enabling policies to acquire physical reasoning skills and recover from unseen conditions. Additionally, emerging research on general-purpose visuomotor foundation models and their combination with reinforcement learning \cite{tri2026, xu2026} presents a promising direction for learning transferable policies across a broad range of dexterous manipulation tasks. 

%% file: bibliography.bib
@article{delgado2019,
   author = {Juan Manuel Davila Delgado and Lukumon Oyedele and Anuoluwapo Ajayi and Lukman Akanbi and Olugbenga Akinade and Muhammad Bilal and Hakeem Owolabi},
   url = {https://doi.org/10.1016/j.jobe.2019.100868}      ,
   journal = {Journal of Building Engineering},
   pages = {100868},
   title = {Robotics and automated systems in construction: Understanding industry-specific challenges for adoption},
   volume = {26},
   year = {2019},
}

@article{wei2023,
  title={Intelligent robots and human-robot collaboration in the construction industry: a review},
  author={Wei, Hsi-Hien and Zhang, Yuting and Sun, Ximing and Chen, Jiayu and Li, Shixian},
  journal={Journal of Intelligent Construction},
  volume={1},
  number={1},
  pages={1--12},
  year={2023},
  publisher={TUP},
url ={https://doi.org/10.26599/JIC.2023.9180002}   
}

@article{laukkanen1999,
   author = {Tuula Laukkanen},
   url = {https://doi.org/10.1080/014461999371826},
   issue = {1},
   journal = {Construction Management and Economics},
   pages = {53-62},
   title = {Construction work and education: occupational health and safety reviewed},
   volume = {17},
   year = {1999},
}

@article{arndt2005,
   author = {V Arndt and D Rothenbacher and U Daniel and B Zschenderlein and S Schuberth and H Brenner},
   url = {https://doi.org/10.1136/oem.2004.018135}  ,
   issue = {8},
   journal = {Occupational and Environmental Medicine},
   pages = {559-566},
   title = {Construction work and risk of occupational disability: a ten year follow up of 14 474 male workers},
   volume = {62},
   year = {2005},
}

@article{bock2015,
   author = {Thomas Bock},
   url = {https://doi.org/10.1016/j.autcon.2015.07.022}    ,
   journal = {Automation in Construction},
   pages = {113-121},
   title = {The future of construction automation: Technological disruption and the upcoming ubiquity of robotics},
   volume = {59},
   year = {2015},
}

@article{musarat2024,
  title={Substitution of workforce with robotics in the construction industry: A wise or witless approach},
  author={Musarat, Muhammad Ali and Alaloul, Wesam Salah and Rostam, Nur Aqilah Qistina Ahmad and Khan, Abdul Mateen},
  journal={Journal of Open Innovation: Technology, Market, and Complexity},
  volume={10},
  number={4},
  pages={100420},
  year={2024},
  publisher={Elsevier},
 url = {https://doi.org/10.1016/j.joitmc.2024.100420}    
}

@article{chen2025,
  title={Advancing robotic assembly in construction: Innovations, challenges, and opportunities},
  author={Chen, Zhengyi and Adel, Arash},
  journal={Automation in Construction},
  volume={178},
  pages={106370},
  year={2025},
  publisher={Elsevier},
  url = {https://doi.org/10.1016/j.autcon.2025.106370} 
}

@article{graser2021,
   author = {Konrad Graser and Arash Adel and Marco Baur and Daniel Sanz Pont and Andreas Thoma},
   url = {https://doi.org/10.1080/24751448.2021.1863668}  ,
   issue = {1},
   journal = {Technology$|$Architecture + Design},
   pages = {38-43},
   title = {Parallel Paths of Inquiry: Detailing for {DFAB HOUSE}},
   volume = {5},
   year = {2021},
}

@article{adel2024,
  title={Feedback-driven adaptive multi-robot timber construction},
  author={Adel, Arash and Ruan, Daniel and McGee, Wesley and Mozaffari, Salma},
  journal={Automation in Construction},
  volume={164},
  pages={105444},
  year={2024},
  publisher={Elsevier},
  url = {https://doi.org/10.1016/j.autcon.2024.105444}   
}

@misc{ruan2026,  
      title={A Latency-Aware Framework for Visuomotor Policy Learning on Industrial Robots}, 
      author={Daniel Ruan and Salma Mozaffari and Sigrid Adriaenssens and Arash Adel},
      year={2026},
      archivePrefix={arXiv},
      url={https://arxiv.org/abs/2602.14255}
}

@InCollection{willmann2016,
  Title = {New Paradigms of the Automatic Robotic Timber Construction in Architecture},
  author = {Jan Willmann and Fabio Gramazio and Matthias Kohler},
  publisher = {Routledge},
  booktitle = {Advancing Wood Architecture: a Computational Approach},
  url = {https://doi.org/10.4324/9781315678825-2} ,
  year = {2017},
  pages = {13--27},
}

@phdthesis{apolinarska2018,
  title={Complex timber structures from simple elements: computational design of novel bar structures for robotic fabrication and assembly},
  author={Apolinarska, Aleksandra Anna},
  year={2018},
  school={ETH Zurich},
  url = {https://doi.org/10.3929/ethz-b-000266723}  ,
}

@inproceedings{adel2018,
   title = {Design of Robotically Fabricated Timber Frame Structures},
   author = {Arash Adel and Andreas Thoma and Matthias Helmreich and Fabio Gramazio and Matthias Kohler},
   city = {Mexico City},
   booktitle = {Recalibration, On Imprecision and Infidelity, Proceedings of the 38th Annual Conference of the Association for Computer Aided Design in Architecture (ACADIA)},
   pages = {394-403},
   year = {2018},
   url = {https://doi.org/10.52842/conf.acadia.2018.394},
   publisher = {{CumInCAD}}
}

@phdthesis{adel2020, 
    author = {Arash Adel},
    title = {Computational Design for Cooperative Robotic Assembly of Nonstandard Timber Frame Buildings},
    school = {ETH Zurich},
    year = {2020},
    url = {https://doi.org/10.3929/ethz-b-000439443},
}

@inproceedings{ruan2023,
   title={Robotic Fabrication of Nail-Laminated Timber: A Case Study Exhibition}, 
   author = {Ruan, Daniel and Adel, Arash},
   booktitle = {Habits of the Anthropocene, Scarcity and Abundance in a Post-Material Economy, Volume 1, Projects Catalog of the 43rd Annual Conference of the Association of Computer Aided Design in Architecture (ACADIA)},
   pages = {220-225},
   year = {2023},
   url = {https://dx.doi.org/10.7302/21575},
   publisher = {{CumInCAD}}
}

@inproceedings{adel2026,
      title={Computational Design and Co-Robotic Fabrication for Material Reuse in Architecture}, 
      author={Arash Adel and Daniel Ruan and Ruxin Xie},
      year={2026},
      booktitle = {COMPUTING for RESILIENCE: Expanding Community Knowledge \& Impact, Proceedings of the 45th Annual Conference of the Association for Computer Aided Design in Architecture (ACADIA)},
      url={https://arxiv.org/abs/2604.24648}, 
}

@article{chai2022,
  title={Computational design and on-site mobile robotic construction of an adaptive reinforcement beam network for cross-laminated timber slab panels},
  author={Chai, Hua and Wagner, Hans Jakob and Guo, Zhixian and Qi, Yue and Menges, Achim and Yuan, Philip F},
  journal={Automation in Construction},
  volume={142},
  pages={104536},
  year={2022},
  publisher={Elsevier},
  url ={https://doi.org/10.1016/j.autcon.2022.104536} 
}

@article{lauer2023,
  title={Automated on-site assembly of timber buildings on the example of a biomimetic shell},
  author={Lauer, Anja Patricia Regina and Benner, Elisabeth and Stark, Tim and Klassen, Sergej and Abolhasani, Sahar and Schroth, Lukas and Gienger, Andreas and Wagner, Hans Jakob and Schwieger, Volker and Menges, Achim and others},
  journal={Automation in Construction},
  volume={156},
  pages={105118},
  year={2023},
  publisher={Elsevier},
 url ={https://doi.org/10.1016/j.autcon.2023.105118}  
}

@inproceedings{leung2021,
  title={Automatic assembly of jointed timber structure using distributed robotic clamps},
  author={Leung, Pok Yin and Apolinarska, Aleksandra Anna and Tanadini, Davide and Gramazio, Fabio and Kohler, Matthias},
  booktitle={{PROJECTIONS}, Proceedings of the 26th International Conference of the Association for Computer-Aided Architectural Design (CAADRIA)},
  volume={1},
  pages={583--592},
  year={2021},
   url = {https://doi.org/10.52842/conf.caadria.2021.1.583},
   publisher = {{CumInCAD}}
}

@article{apolinarska2021,
  title={Robotic assembly of timber joints using reinforcement learning},
  author={Apolinarska, Aleksandra Anna and Pacher, Matteo and Li, Hui and Cote, Nicholas and Pastrana, Rafael and Gramazio, Fabio and Kohler, Matthias},
  journal={Automation in Construction},
  volume={125},
  pages={103569},
  year={2021},
  publisher={Elsevier},
  url = {https://doi.org/10.1016/j.autcon.2021.103569}
}

@article{kramberger2022,
  title={Robotic assembly of timber structures in a human-robot collaboration setup},
  author={Kramberger, Aljaz and Kunic, Anja and Iturrate, I{\~n}igo and Sloth, Christoffer and Naboni, Roberto and Schlette, Christian},
  journal={Frontiers in Robotics and AI},
  volume={8},
  pages={768038},
  year={2022},
  publisher={Frontiers Media SA},
  url = {https://doi.org/10.3389/frobt.2021.768038}
}

@article{duan2024,
  title={Training of construction robots using imitation learning and environmental rewards},
  author={Duan, Kangkang and Zou, Zhengbo and Yang, TY},
  journal={Computer-Aided Civil and Infrastructure Engineering},
  volume={40},
  number={9},
  pages={1150--1165},
  year={2024},
  publisher={Wiley Online Library},
  url={https://doi.org/10.1111/mice.13394}
}

@article{li2023,   
  title={Enhancing construction robot learning for collaborative and long-horizon tasks using generative adversarial imitation learning},
  author={Li, Rui and Zou, Zhengbo},
  journal={Advanced Engineering Informatics},
  volume={58},
  pages={102140},
  year={2023},
  publisher={Elsevier},
  url={https://doi.org/10.1016/j.aei.2023.102140}
}

@article{duan2025,
  title={Visual--tactile learning of robotic cable-in-duct installation skills},
  author={Duan, Boyi and Qian, Kun and Liu, Aohua and Luo, Shan},
  journal={Automation in Construction},
  volume={170},
  pages={105905},
  year={2025},
  publisher={Elsevier},
  url= {https://doi.org/10.1016/j.autcon.2024.105905}
}

@article{yu2024,
  title={Cloud-based hierarchical imitation learning for scalable transfer of construction skills from human workers to assisting robots},
  author={Yu, Hongrui and Kamat, Vineet R and Menassa, Carol C},
  journal={Journal of Computing in Civil Engineering},
  volume={38},
  number={4},
  pages={04024019},
  year={2024},
  publisher={American Society of Civil Engineers},
  url={https://doi.org/10.1061/JCCEE5.CPENG-5731}
}

@article{sun2026,
  title={Mobile robotic rebar cage assembly via imitation learning},
  author={Sun, Tao and Han, Beining and Wu, Jimmy and Rusinkiewicz, Szymon and Shao, Yi},
  journal={Automation in Construction},
  volume={181},
  pages={106671},
  year={2026},
  publisher={Elsevier},
  url ={https://doi.org/10.1016/j.autcon.2025.106671}
}

@article{huang2026,
title = {Act or ask: Interactive construction robots via vision–language models with confidence-guided decision deferral},
author = {Lei Huang and Zhengbo Zou},
journal = {Advanced Engineering Informatics},
volume = {72},
pages = {104454},
year = {2026},
issn = {1474-0346},
url = {https://doi.org/10.1016/j.aei.2026.104454},
}

@article{delgado2022,
  title={Robotics in construction: A critical review of the reinforcement learning and imitation learning paradigms},
  author={Delgado, Juan Manuel Davila and Oyedele, Lukumon},
  journal={Advanced Engineering Informatics},
  volume={54},
  pages={101787},
  year={2022},
  publisher={Elsevier},
  url={https://doi.org/10.1016/j.aei.2022.101787}
}

@article{zhou2026, 
  title={Robust robotic assembly via hierarchical diffusion policy-guided reinforcement learning},
  author={Zhou, Yibang and Li, Xiangkai and Yin, Yue and Chen, Liwei and Xu, Haiming and Fu, Jiajie and Zhou, Aoyu and Yi, Jianjun},
  journal={Advanced Engineering Informatics},
  volume={71},
  pages={104399},
  year={2026},
  publisher={Elsevier},
  url = {https://doi.org/10.1016/j.aei.2026.104399}
}

@article{kommey2025,
  title={A Compact Review of Industrial Robots: Dynamic Modeling, Control Strategies, and Operational Challenges},
  author={Kommey, Benjamin and Essah, Stephanie and Kuusofaa, David Dery and Jnr, Samuel Boahene},
  journal={Andalas Journal of Electrical and Electronic Engineering Technology},
  volume={5},
  number={2},
  pages={50--64},
  year={2025},
  url={https://doi.org/10.25077/ajeeet.v5i2.165}
}

@article{yang2024,
  title={Challenges and potential for human--robot collaboration in timber prefabrication},
  author={Yang, Xiliu and Amtsberg, Felix and Sedlmair, Michael and Menges, Achim},
  journal={Automation in Construction},
  volume={160},
  pages={105333},
  year={2024},
  publisher={Elsevier},
  url={https://doi.org/10.1016/j.autcon.2024.105333}  
}

@book{benson1981,
  title={Building the timber frame house: The revival of a forgotten craft},
  author={Benson, Tedd},
  year={1981},
  publisher={Simon and Schuster}
}

@article{fang2023,
  title={Mortise-and-tenon joinery for modern timber construction: Quantifying the embodied carbon of an alternative structural connection},
  author={Fang, Demi and Mueller, Caitlin},
  journal={Architecture, Structures and Construction},
  volume={3},
  number={1},
  pages={11--24},
  year={2023},
  publisher={Springer},
  url = {https://doi.org/10.1007/s44150-021-00018-5}
}

@article{levine2016,
  title={End-to-end training of deep visuomotor policies},
  author={Levine, Sergey and Finn, Chelsea and Darrell, Trevor and Abbeel, Pieter},
  journal={Journal of Machine Learning Research},
  volume={17},
  number={39},
  pages={1--40},
  year={2016},
  url={http://jmlr.org/papers/v17/15-522.html}
}

@inproceedings{finn2017,
  title={One-shot visual imitation learning via meta-learning},
  author={Finn, Chelsea and Yu, Tianhe and Zhang, Tianhao and Abbeel, Pieter and Levine, Sergey},
  booktitle={Proceedings of the Conference on Robot Learning {CoRL}},
  pages={357--368},
  year={2017},
  organization={PMLR},
  url={https://proceedings.mlr.press/v78/finn17a.html}
}

@article{levine2018,
  title={Learning hand-eye coordination for robotic grasping with deep learning and large-scale data collection},
  author={Levine, Sergey and Pastor, Peter and Krizhevsky, Alex and Ibarz, Julian and Quillen, Deirdre},
  journal={The International journal of robotics research},
  volume={37},
  number={4-5},
  pages={421--436},
  year={2018},
  publisher={SAGE Publications Sage UK: London, England},
  url={https://doi.org/10.1177/0278364917710318}
}

@inproceedings{kalashnikov2018,
  title={Scalable deep reinforcement learning for vision-based robotic manipulation},
  author={Kalashnikov, Dmitry and Irpan, Alex and Pastor, Peter and Ibarz, Julian and Herzog, Alexander and Jang, Eric and Quillen, Deirdre and Holly, Ethan and Kalakrishnan, Mrinal and Vanhoucke, Vincent and others},
  booktitle={Proccedings of Conference on robot learning (CoRL)},
  pages={651--673},
  volume = {87},
  year={2018},
  organization={PMLR},
  url={https://proceedings.mlr.press/v87/kalashnikov18a.html},
}

@misc{brohan2023,
      title={{RT-1}: Robotics Transformer for Real-World Control at Scale}, 
      author={Anthony Brohan and Noah Brown and Justice Carbajal and Yevgen Chebotar and Joseph Dabis and Chelsea Finn and Keerthana Gopalakrishnan and Karol Hausman and Alex Herzog and Jasmine Hsu and Julian Ibarz and Brian Ichter and Alex Irpan and Tomas Jackson and Sally Jesmonth and Nikhil J Joshi and Ryan Julian and Dmitry Kalashnikov and Yuheng Kuang and Isabel Leal and Kuang-Huei Lee and Sergey Levine and others},
      year={2023},
      eprint={2212.06817},
      archivePrefix={arXiv},
      primaryClass={cs.RO},
      url={https://arxiv.org/abs/2212.06817}, 
}

@article{chi2024a,
  title={Diffusion policy: Visuomotor policy learning via action diffusion},
  author={Chi, Cheng and Xu, Zhenjia and Feng, Siyuan and Cousineau, Eric and Du, Yilun and Burchfiel, Benjamin and Tedrake, Russ and Song, Shuran},
  journal={The International Journal of Robotics Research},
  volume={44},
  number={10-11},
  pages={1684--1704},
  year={2024},
  url={https://doi.org/10.1177/02783649241273668}
}

@misc{chi2024b,
      title={Universal Manipulation Interface: In-The-Wild Robot Teaching Without In-The-Wild Robots}, 
      author={Cheng Chi and Zhenjia Xu and Chuer Pan and Eric Cousineau and Benjamin Burchfiel and Siyuan Feng and Russ Tedrake and Shuran Song},
      year={2024},
      eprint={2402.10329},
      archivePrefix={arXiv},
      primaryClass={cs.RO},
      url={https://arxiv.org/abs/2402.10329}, 
}

@inproceedings{hou2025,
  title={Adaptive compliance policy: Learning approximate compliance for diffusion guided control},
  author={Hou, Yifan and Liu, Zeyi and Chi, Cheng and Cousineau, Eric and Kuppuswamy, Naveen and Feng, Siyuan and Burchfiel, Benjamin and Song, Shuran},
  booktitle={2025 IEEE International Conference on Robotics and Automation (ICRA)},
  pages={4829--4836},
  year={2025},
  organization={IEEE},
  url = {https://doi.org/10.1109/ICRA55743.2025.11128452}
}

@misc{yang2025,
      title={Physics-Driven Data Generation for Contact-Rich Manipulation via Trajectory Optimization}, 
      author={Lujie Yang and H. J. Terry Suh and Tong Zhao and Bernhard Paus Graesdal and Tarik Kelestemur and Jiuguang Wang and Tao Pang and Russ Tedrake},
      year={2025},
      eprint={2502.20382},
      archivePrefix={arXiv},
      url={https://arxiv.org/abs/2502.20382}, 
}

@article{huaijiang2025,
  author={Zhu, Huaijiang and Zhao, Tong and Ni, Xinpei and Wang, Jiuguang and Fang, Kuan and Righetti, Ludovic and Pang, Tao},
  journal={IEEE Robotics and Automation Letters}, 
  title={Should We Learn Contact-Rich Manipulation Policies From Sampling-Based Planners?}, 
  year={2025},
  volume={10},
  number={6},
  pages={6248-6255},
  url={https://doi.org/10.1109/LRA.2025.3564701}
}

@misc{black2026,
      title={$\pi_0$: A Vision-Language-Action Flow Model for General Robot Control}, 
      author={Kevin Black and Noah Brown and Danny Driess and Adnan Esmail and Michael Equi and Chelsea Finn and Niccolo Fusai and Lachy Groom and Karol Hausman and Brian Ichter and Szymon Jakubczak and Tim Jones and Liyiming Ke and Sergey Levine and Adrian Li-Bell and Mohith Mothukuri and Suraj Nair and Karl Pertsch and Lucy Xiaoyang Shi and James Tanner and Quan Vuong and Anna Walling and Haohuan Wang and Ury Zhilinsky},
      year={2026},
      eprint={2410.24164},
      archivePrefix={arXiv},
      primaryClass={cs.LG},
      url={https://arxiv.org/abs/2410.24164}, 
}

@article{stadelmann2019,
   author = {Lukas Stadelmann and Timothy Sandy and Andreas Thoma and Jonas Buchli},
   url = {https://doi.org/10.1109/LRA.2019.2891499}  ,
   issue = {2},
   journal = {IEEE Robotics and Automation Letters},
   pages = {546-553},
   title = {End-Effector Pose Correction for Versatile Large-Scale Multi-Robotic Systems},
   volume = {4},
   year = {2019},
}

@inproceedings{gandia2022,
   author = {Augusto Gandia and Fabio Gramazio and Matthias Kohler},
   pages = {4-23},
   title = {Tolerance-aware Design of Robotically Assembled Spatial Structures},
   year = {2022},
   booktitle = {Hybrids \& Haecceities, Proceedings of the 42nd Annual Conference of the Association for Computer Aided Design in Architecture (ACADIA)},
   url ={https://papers.cumincad.org/cgi-bin/works/Show?acadia22_4},
   publisher = {{CumInCAD}}
}

@inproceedings{helm2016,
   author = {Volker Helm and Michael Knauss and Thomas Kohlhammer and Fabio Gramazio and Matthias Kohler},
   city = {New York: Routledge, 2016.},
   url = {https://doi.org/10.4324/9781315678825-3} ,
   booktitle = {Advancing Wood Architecture: A Computational Approach},
   pages = {29-44},
   publisher = {Routledge},
   title = {Additive robotic fabrication of complex timber structures},
   year = {2016},
}

@article{cote2024,
  title={Adaptive robotic construction of wood frames},
  author={Cote, Nicholas and Tish, Daniel and Koehle, Michael and Koga, Yotto and Chitta, Sachin},
  journal={Construction Robotics},
  volume={8},
  number={1},
  pages={8},
  year={2024},
  publisher={Springer},
  url ={https://doi.org/10.1007/s41693-024-00122-0}   
}

@article{xie2026,
  title={Advancing robotic automation in wood-framed construction using vision-driven adaptive control},
  author={Xie, Chao and Alwisy, Aladdin},
  journal={Automation in Construction},
  volume={185},
  pages={106858},
  year={2026},
  publisher={Elsevier},
  url = {https://doi.org/10.1016/j.autcon.2026.106858}
}

@inproceedings{helmreich2022,
   author={Helmreich, MATTHIAS and Mayer, HANNES and Pacher, M and Nakajima, T and Kuroki, M and Tsubata, S and Gramazio, F and Kohler, M},
   url = {https://doi.org/10.52842/conf.caadria.2022.2.111},
   pages = {111--120},
   title = {Robotic Assembly of Modular Multi-Storey Timber-Only Frame Structures Using Traditional Wood Joinery},
   booktitle={Proceedings of the 27th International Conference for the Association for Computer-Aided Architectural Design Research in Asia (CAADRIA)},
   year = {2022},
   publisher = {{CumInCAD}}
}

@article{albu2007,
  title={A unified passivity-based control framework for position, torque and impedance control of flexible joint robots},
  author={Albu-Sch{\"a}ffer, Alin and Ott, Christian and Hirzinger, Gerd},
  journal={The international journal of robotics research},
  volume={26},
  number={1},
  pages={23--39},
  year={2007},
  publisher={Sage Publications Sage CA: Thousand Oaks, CA},
url = {https://doi.org/10.1177/0278364907073776}
}

@article{suarez2018,
  title={Can robots assemble an IKEA chair?},
  author={Su{\'a}rez-Ruiz, Francisco and Zhou, Xian and Pham, Quang-Cuong},
  journal={Science Robotics},
  volume={3},
  number={17},
  pages={eaat6385},
  year={2018},
  publisher={American Association for the Advancement of Science},
  url = {https://doi.org/10.1126/scirobotics.aat6385}
}

@inproceedings{vecerik2019,
  title={A practical approach to insertion with variable socket position using deep reinforcement learning},
  author={Vecerik, Mel and Sushkov, Oleg and Barker, David and Roth{\"o}rl, Thomas and Hester, Todd and Scholz, Jon},
  booktitle={2019 international conference on robotics and automation (ICRA)},
  pages={754--760},
  year={2019},
  organization={IEEE},
  url = {https://doi.org/10.1109/ICRA.2019.8794074}
}

@inproceedings{schoettler2020,
  title={Deep reinforcement learning for industrial insertion tasks with visual inputs and natural rewards},
  author={Schoettler, Gerrit and Nair, Ashvin and Luo, Jianlan and Bahl, Shikhar and Ojea, Juan Aparicio and Solowjow, Eugen and Levine, Sergey},
  booktitle={Proceedingd of 2020 IEEE/RSJ International Conference on Intelligent Robots and Systems (IROS)},
  pages={5548--5555},
  year={2020},
  organization={IEEE},
  url={https://doi.org/10.1109/IROS45743.2020.9341714}
}

@inproceedings{johannsmeier2019,
  title={A framework for robot manipulation: Skill formalism, meta learning and adaptive control},
  author={Johannsmeier, Lars and Gerchow, Malkin and Haddadin, Sami},
  booktitle={{International Conference on Robotics and Automation (ICRA)}},
  pages={5844--5850},
  year={2019},
  organization={IEEE},
  url = {https://doi.org/10.1109/ICRA.2019.8793542}
}

@inproceedings{robeller2017,
  title={Robotic integral attachment},
  author={Robeller, Christopher and Weinand, Yves and Helm, Volker and Thoma, Andreas and Gramazio, Fabio and Kohler, Matthias},
  booktitle={Proceedings of Fabricate 2017: Rethinking Design and Construction},
  volume={3},
  pages={92--97},
  year={2017},
  publisher={UCL Press},
  url = {https://doi.org/10.2307/j.ctt1n7qkg7.16}
}

@phdthesis{rogeau2023,
  title={Robotic Assembly of Integrally-Attached Timber Plate Structures: From Computational Design to Automated Construction},
  author={Rogeau, Nicolas Henry Pierre Louis},
  year={2023},
  school={EPFL},
  url = {https://infoscience.epfl.ch/entities/publication/6fd77403-f912-4f03-a68c-18a3bac91960}
}

@inproceedings{seo2023,
  author={Seo, Mingyo and Han, Steve and Sim, Kyutae and Bang, Seung Hyeon and Gonzalez, Carlos and Sentis, Luis and Zhu, Yuke},
  booktitle={2023 IEEE-RAS 22nd International Conference on Humanoid Robots (Humanoids)}, 
  title={Deep Imitation Learning for Humanoid Loco-manipulation Through Human Teleoperation}, 
  year={2023},
  pages={1-8},
  url ={https://doi.org/10.1109/Humanoids57100.2023.10375203}
}

@inproceedings{wang2023,
  title     = {{MimicPlay}: Long-Horizon Imitation Learning by Watching Human Play},
  author    = {Chen Wang and Linxi Fan and Jiankai Sun and Ruohan Zhang and Li Fei-Fei and Danfei Xu and Yuke Zhu and Anima Anandkumar},
  booktitle = {Proceedings of The 7th Conference on Robot Learning (CoRL)},
  volume    = {229},
  pages     = {201--221},
  year      = {2023},
 url ={https://doi.org/10.48550/arXiv.2302.12422}
}

@inproceedings{shaw2023,
  title = 	 {{VideoDex}: Learning Dexterity from Internet Videos},
  author =       {Shaw, Kenneth and Bahl, Shikhar and Pathak, Deepak},
  booktitle = 	 {Proceedings of The 6th Conference on Robot Learning},
  pages = 	 {654--665},
  year = 	 {2023},
  editor = 	 {Liu, Karen and Kulic, Dana and Ichnowski, Jeff},
  volume = 	 {205},
  series = 	 {Proceedings of Machine Learning Research},
  publisher =    {PMLR},
  url = 	 {https://proceedings.mlr.press/v205/shaw23a.html}
}

@inproceedings{zhao2023,
  title     = {Learning Fine-Grained Bimanual Manipulation with Low-Cost Hardware},
  author    = {T. Z. Zhao and V. Kumar and S. Levine and C. Finn},
  booktitle = {Proceedings of Robotics: Science and Systems XIX},
  year      = {2023},
  organization = {Robotics: Science and Systems Foundation},
  url       = {https://www.roboticsproceedings.org/rss19/p016.pdf}
}

@InProceedings{zhao2025,
  title = 	 {{ALOHA} Unleashed: A Simple Recipe for Robot Dexterity},
  author =       {Zhao, Tony Z. and Tompson, Jonathan and Driess, Danny and Florence, Pete and Ghasemipour, Seyed Kamyar Seyed and Finn, Chelsea and Wahid, Ayzaan},
  booktitle = 	 {Proceedings of The 8th Conference on Robot Learning},
  pages = 	 {1910--1924},
  year = 	 {2025},
  editor = 	 {Agrawal, Pulkit and Kroemer, Oliver and Burgard, Wolfram},
  volume = 	 {270},
  series = 	 {Proceedings of Machine Learning Research},
  month = 	 {06--09 Nov},
  publisher =    {PMLR},
  url = 	 {https://proceedings.mlr.press/v270/zhao25b.html},
}

@inproceedings{ze2024,
  author = {Ze, Y. and Zhang, G. and Zhang, K. and Hu, C. and Wang, M. and Xu, H.},
  title = {{3D} Diffusion Policy: Generalizable Visuomotor Policy Learning via Simple {3D} Representations},
  booktitle = {ICRA 2024 Workshop on 3D Visual Representations for Robot Manipulation},
  year = {2024},
  url = {https://www.roboticsproceedings.org/rss20/p067.pdf}
}

@inproceedings{ho2020,
  title={Denoising diffusion probabilistic models},
  author={Ho, Jonathan and Jain, Ajay and Abbeel, Pieter},
  booktitle={Proceedings of the 34th International Conference on Neural Information Processing Systems (NeurIPS)},
  volume={33},
  pages={6840--6851},
  year={2020},
   url = {https://doi.org/10.48550/arXiv.2006.11239},
}

@misc{song2022,
  title={Denoising Diffusion Implicit Models}, 
  author={Jiaming Song and Chenlin Meng and Stefano Ermon},
  year={2022},
  eprint={2010.02502},
  archivePrefix={arXiv},
  primaryClass={cs.LG},
  url={https://doi.org/10.48550/arXiv.2010.02502}, 
}

@article{florence2020,
  title={Self-supervised correspondence in visuomotor policy learning},
  author={Florence, Peter and Manuelli, Lucas and Tedrake, Russ},
  journal={IEEE Robotics and Automation Letters},
  volume={5},
  number={2},
  pages={492--499},
  year={2020},
  publisher={IEEE},
 url = {https://doi.org/10.1109/LRA.2019.2956365}
}

@inproceedings{shafiullah2022,
 author = {Shafiullah, Nur Muhammad and Cui, Zichen and Altanzaya, Ariuntuya (Arty) and Pinto, Lerrel},
  booktitle={Proceedings of Advances in Neural Information Processing Systems (NeurIPS)},
 editor = {S. Koyejo and S. Mohamed and A. Agarwal and D. Belgrave and K. Cho and A. Oh},
 pages = {22955--22968},
 publisher = {Curran Associates, Inc.},
 title = {Behavior Transformers: Cloning k modes with one stone},
 url = {https://proceedings.neurips.cc/paper_files/paper/2022/file/90d17e882adbdda42349db6f50123817-Paper-Conference.pdf},
 volume = {35},
 year = {2022}
}

@inproceedings{florence2022,
  title={Implicit behavioral cloning},
  author={Florence, Pete and Lynch, Corey and Zeng, Andy and Ramirez, Oscar A and Wahid, Ayzaan and Downs, Laura and Wong, Adrian and Lee, Johnny and Mordatch, Igor and Tompson, Jonathan},
  booktitle={Proceedings of the Conference on Robot Learning (CoRL)},
  pages={158--168},
  year={2022},
  organization={PMLR},
  url = {https://proceedings.mlr.press/v164/florence22a.html},
}

@inproceedings{jarrett2020,
  title={Strictly batch imitation learning by energy-based distribution matching},
  author={Jarrett, Daniel and Bica, Ioana and van der Schaar, Mihaela},
  booktitle={Proceedings of Advances in Neural Information Processing Systems (NeurIPS)},
  volume={33},
  pages={7354--7365},
  year={2020},
  url= {https://proceedings.neurips.cc/paper_files/paper/2020/hash/524f141e189d2a00968c3d48cadd4159-Abstract.html}
}

@misc{wu2025,
      title={{TacDiffusion}: Force-domain Diffusion Policy for Precise Tactile Manipulation}, 
      author={Yansong Wu and Zongxie Chen and Fan Wu and Lingyun Chen and Liding Zhang and Zhenshan Bing and Abdalla Swikir and Sami Haddadin and Alois Knoll},
      year={2025},
      eprint={2409.11047},
      archivePrefix={arXiv},
      primaryClass={cs.RO},
      url={https://arxiv.org/abs/2409.11047}, 
}

@misc{abb,
  author = {{ABB Group}},
  title = {{IRB 4600 40kg/2,55m}},
  url = {https://new.abb.com/products/robotics/robots/articulated-robots/irb-4600},
  note  = {{{A}ccessed March 2026}}
}

@misc{ati,
  author = {{ATI Industrial Automation}},
  title = {{F/T Sensor: Delta IP60}},
  url = {https://www.ati-ia.com/products/ft/ft_models.aspx?id=delta+ip60},
  note  = {{{A}ccessed March 2026}}
}

@misc{schunk-ac,
  author = {{Schunk}},
  title = {{OPR 081-P00 Anti-collision and overload protection sensor}},
  url = {https://schunk.com/us/en/automation-technology/anti-collision-unit/opr/c/PGR_1105},
  note  = {{A}ccessed March 2026}
}

@manual{rapid2017,
  title        = {Technical Reference Manual - RAPID Instructions, Functions and Data Types},
  author       = {{ABB Robotics}},
  year         = {2017},
  note         = {{RobotWare} 6.05},
  organization = {ABB AB, Robotics and Motion},
  address      = {Västerås, Sweden},
}

@misc{beckoff,
  author = {{Beckoff}},
  title = {{CX2062 \textbar{} Embedded PC with Intel\textsuperscript{\textregistered} Xeon\textsuperscript{\textregistered} D-1548}},
  url = {https://www.beckhoff.com/en-us/products/ipc/embedded-pcs/cx20x2-intel-r-xeon-r-d/cx2062.html},
  note = {{A}ccessed March 2026}
}

@misc{ros2,
  author = {{Open Robotics}},
  title = {{ROS 2 Jazzy Jalisco}},
  year = {2023},
  url = {https://docs.ros.org/en/jazzy/},
  note = {{A}ccessed March 2026}
}

@misc{htc,
  author = {{HTC Corporation}},
  title = {{HTC VIVE Pro 2}},
  url = {https://www.vive.com/us/product/vive-pro2/overview/},
  note  = {{A}ccessed March 2026}
  }

@misc{openvr,
  author = {{Valve Corporation}},
  title = {{OpenVR SDK}},
  year = {2024},
  url = {https://github.com/ValveSoftware/openvr},
  note = {{A}ccessed March 2026}
}

@article{Butterworth1930,
  title={On the theory of filter amplifiers},
  author={Butterworth, Stephen},
  journal={Wireless Engineer},
  volume={7},
  number={6},
  pages={536--541},
  year={1930}
}

@misc{structurecraft,
  author       = {{StructureCraft}},
  title        = {Structural Engineers \& Mass Timber Builders},
  url          = {https://structurecraft.com/},
  note  = {{A}ccessed March 2026}
}

@misc{shinohara,
  author       = {{Shinohara Shoten Co., Ltd.}},
  title        = {The specialist group for timber construction},
  url          = {https://en.shinoharashoten.com/shinohara},
  note  = {{A}ccessed March 2026}
}

@misc{bogert2024,
  title={{Built Different}: Tactile Perception to Overcome Cross-Embodiment Capability Differences in Collaborative Manipulation}, 
  author={William van den Bogert and Madhavan Iyengar and Nima Fazeli},
  year={2024},
  eprint={2409.14896},
  archivePrefix={arXiv},
  primaryClass={cs.RO},
  url={https://arxiv.org/abs/2409.14896}, 
}

@inproceedings{zhou2019,
  title={On the continuity of rotation representations in neural networks},
  author={Zhou, Yi and Barnes, Connelly and Lu, Jingwan and Yang, Jimei and Li, Hao},
  booktitle={Proceedings of the IEEE/CVF conference on computer vision and pattern recognition},
  pages={5745--5753},
  year={2019},
  url = {https://openaccess.thecvf.com/content_CVPR_2019/papers/Zhou_On_the_Continuity_of_Rotation_Representations_in_Neural_Networks_CVPR_2019_paper.pdf}
}

@inproceedings{nichol2021,
  title = {Improved Denoising Diffusion Probabilistic Models},
  author = {Nichol, Alexander Quinn and Dhariwal, Prafulla},
  booktitle = {Proceedings of the 38th International Conference on Machine Learning},
  pages = {8162--8171},
  year = {2021},
  volume = {139},
  series = {Proceedings of Machine Learning Research},
  month = {18--24 Jul},
  publisher = {PMLR},
  url = {https://proceedings.mlr.press/v139/nichol21a.html}
}

@inproceedings{craney2020,
   title = {Engrained Performance: Performance-Driven Computational Design of a Robotically Assembled Shingle Facade System},
   author = {Ryan Craney and Arash Adel},
   booktitle = {Distributed Proximities, Proceedings of the 40th Annual Conference of the Association of Computer Aided Design in Architecture (ACADIA)},
   pages = {604-613},
   year = {2020},
   url = {https://doi.org/10.52842/conf.acadia.2020.1.604} ,
   publisher = {{CumInCAD}}
}

@article{tri2026,
  title={A careful examination of large behavior models for multitask dexterous manipulation},
  author={Barreiros, Jose and Beaulieu, Andrew and Bhat, Aditya and Cory, Rick and Cousineau, Eric and Dai, Hongkai and Fang, Ching-Hsin and Hashimoto, Kunimatsu and Irshad, Muhammad Zubair and Itkina, Masha and others},
  journal={Science Robotics},
  volume={11},
  number={113},
  pages={eaea6201},
  year={2026},
  publisher={American Association for the Advancement of Science}
}

@misc{gr2025,
  title={Gemini {R}obotics 1.5: Pushing the frontier of generalist robots with advanced embodied reasoning, thinking, and motion transfer},
  author={{Gemini Robotics Team} and Abdolmaleki, Abbas and Abeyruwan, Saminda and Ainslie, Joshua and Alayrac, Jean-Baptiste and Arenas, Montserrat Gonzalez and Balakrishna, Ashwin and Batchelor, Nathan and Bewley, Alex and Bingham, Jeff and others},
  year={2025},
  eprint={2510.03342},
  url={https://arxiv.org/abs/2510.03342}, 
}

@misc{xu2026,
      title={{RL} {T}oken: Bootstrapping Online {RL} with Vision-Language-Action Models}, 
      author={Charles Xu and Jost Tobias Springenberg and Michael Equi and Ali Amin and Adnan Esmail and Sergey Levine and Liyiming Ke},
      year={2026},
      eprint={2604.23073},
      archivePrefix={arXiv},
      primaryClass={cs.LG},
      url={https://arxiv.org/abs/2604.23073}, 
}

@incollection{mccullagh1989,
  title={Binary data},
  author={McCullagh, Peter and Nelder, John A},
  booktitle={Generalized linear models},
  pages={98--148},
  year={1989},
  publisher={Springer},
  url={https://doi.org/10.1007/978-3-642-60232-0_7},
}

@article{holm1979,
  title={A simple sequentially rejective multiple test procedure},
  author={Holm, Sture},
  journal={Scandinavian journal of statistics},
  pages={65--70},
  year={1979},
  publisher={JSTOR},
  url={https://www.jstor.org/stable/4615733},
}
